\documentclass[a4paper,fleqn]{cas-sc}

\usepackage[authoryear,longnamesfirst]{natbib}

\usepackage{tikz-dependency}
\usepackage{tikz}
\usepackage{tikz-qtree}
\usepackage{latexsym}
\usepackage{amssymb}
\usepackage{amsmath}
\usepackage{arydshln}

\newcommand{\new}[1]{\textcolor{black}{#1}}
\newcommand{\added}[1]{\textcolor{black}{#1}}

\def\tsc#1{\csdef{#1}{\textsc{\lowercase{#1}}\xspace}}
\tsc{WGM}
\tsc{QE}

\begin{document}
\let\WriteBookmarks\relax
\def\floatpagepagefraction{1}
\def\textpagefraction{.001}

    \shorttitle{Discontinuous Grammar as a Foreign Language}

\shortauthors{D. Fern\'andez-Gonz\'alez, C. G\'omez-Rodr\'iguez.}  

\title [mode = title]{Discontinuous Grammar as a Foreign Language}



%

\author[1]{Daniel Fern\'{a}ndez-Gonz\'{a}lez}[orcid=0000-0002-6733-2371]

\cormark[1]


\ead{d.fgonzalez@udc.es}

\ead[url]{https://danifg.github.io}

\credit{Conceptualization, methodology, software, validation, formal analysis, investigation, data curation, writing - original draft, writing - review \& editing, visualization}

\affiliation[1]{organization={Universidade da Coru\~{n}a, CITIC, FASTPARSE Lab, LyS Group, Depto. de Ciencias de la Computaci\'{o}n y Tecnolog\'{i}as de la Informaci\'{o}n},
            addressline={Campus de Elvi\~{n}a, s/n }, 
            city={A Coru\~{n}a},
            postcode={15071}, 
            country={Spain}}

\author[1]{Carlos G\'{o}mez-Rodr\'{i}guez}[orcid=0000-0003-0752-8812]


\ead{carlos.gomez@udc.es}

\ead[url]{http://www.grupolys.org/~cgomezr/}

\credit{Validation, formal analysis, writing - review \& editing, supervision, project administration, funding acquisition}


\cortext[1]{Corresponding author}



\begin{abstract}
In order to 
achieve deep natural language understanding,
syntactic constituent parsing is a vital step,
highly demanded 
by many artificial intelligence systems to process both text and speech. 
One of the most recent proposals is the use of 
standard
sequence-to-sequence models to perform constituent parsing as a machine translation task, instead of applying task-specific parsers. While they show a competitive performance, these text-to-parse transducers are still lagging behind classic techniques in terms of accuracy, coverage and speed. To close the gap, we 
here
extend the framework of sequence-to-sequence models for constituent parsing, not only by providing a more powerful neural architecture for improving their performance,
but also by enlarging their coverage to handle the most complex syntactic phenomena: discontinuous structures. To that end, we design several novel linearizations that can fully produce discontinuities and, for the first time, we test a sequence-to-sequence model on the main discontinuous benchmarks, obtaining competitive results on par with task-specific
discontinuous constituent 
parsers and achieving state-of-the-art scores on the (discontinuous) English Penn Treebank.

\end{abstract}



\maketitle

\section{Introduction}
Syntactic parsing is a fundamental problem for Natural Language Processing in its pursuit towards deep understanding and computer-friendly representation of human 
linguistic input. Parsers are in charge of efficiently and accurately providing syntactic information so that it can be used for downstream artificial intelligence applications such as machine translation \citep{zhang-etal-2019-syntax-enhanced,YANG2020105042,ZHANG2021103427}, opinion mining \citep{zhang-etal-2020-syntax}, relation and event extraction \citep{Nguyen2019}, question answering \citep{CAOPMID:31562071}, summarization \citep{balachandran-etal-2021-structsum}, sentiment classification \citep{bai-etal-2021-syntax}, sentence classification \citep{ZHANG2021103427} or semantic role labeling and named entity recognition \citep{sachan-etal-2021-syntax}, among others.

\begin{figure}
\begin{center}
\includegraphics[width=0.9\columnwidth]{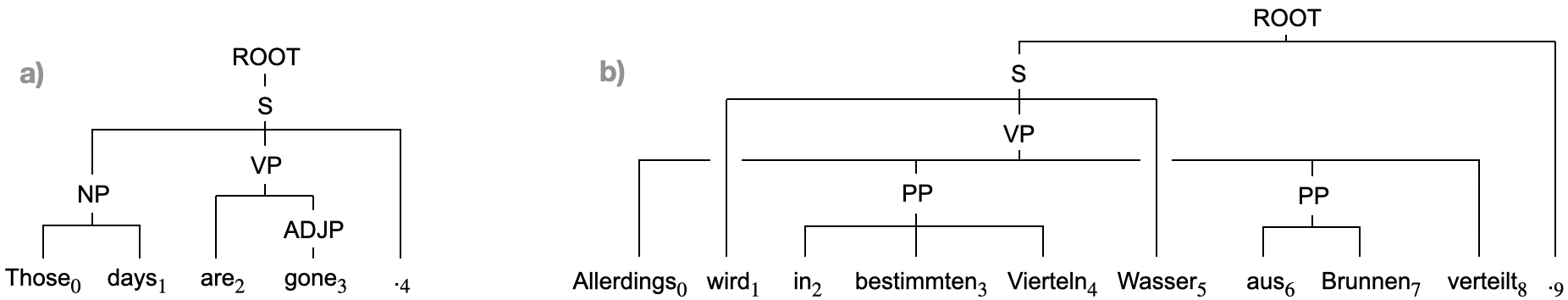}
\end{center}
\caption{Continuous (a) and discontinuous (b) constituent trees taken from PTB train and NEGRA dev splits, respectively.}
\label{fig:trees}
\end{figure}

One of the widely-used formalisms for representing the grammatical structure of a given sentence in human languages is \textit{constituent trees}. These structures decompose the sentence into \textit{constituents} 
(also called \textit{phrases}) 
and establish hierarchical relations between them and the sentence's words, 
resulting in
a tree structure. While regular (or \textit{continuous}) constituent trees (as the one depicted in Figure~\ref{fig:trees}(a)) are enough for representing a wide range of syntactic structures, it is necessary to use \textit{discontinuous} constituent trees to fully describe all linguistic phenomena present in human languages \citep{gebhardt-etal-2017-hybrid}. Although producing the latter is considered 
an especially challenging problem
in constituent parsing \citep{Corro2020SpanbasedDC}, 
they are necessary for adequately representing some syntactic phenomena that occur in almost the 20\% of the sentences from the most widely-used syntactically-annotated corpus of English, the Penn Treebank \citep{marcus93} such as cross-serial dependencies, dislocations, long-distance extractions and some \new{wh-movements \citep{evang-kallmeyer-2011-plcfrs}}, which require constituents with discontinuous spans and result in phrase structure trees with crossing branches. 
\added{For instance, it can be seen in Figure~\ref{fig:trees}(b) that the span of the constituent \textit{VP} (composed of the words \textit{Allerdings}, \textit{in}, \textit{bestimmten}, \textit{Vierteln}, \textit{aus}, \textit{Brunnen} and \textit{verteilt}) is a discontinuous string, since it is interrupted by the words $\textit{wird}$ and $\textit{Wasser}$ from constituent \textit{S}. Unlike the constinuous constituent tree in Figure~\ref{fig:trees}(a)), this phenomenon generates a discontinuous phrase structure tree with two crossing branches.}

In the last three decades, different techniques have been proposed for performing continuous and discontinuous constituent parsing.
One of the most recent approaches, introduced by \citet{Vinyals2015}, consists in using generic \textit{sequence-to-sequence} models to directly \textit{translate} 
text into phrase structure trees, mimicking a machine translation task where these models had previously achieved considerable success \citep{Sutskever2014}. This 
made it possible to perform continuous constituent parsing,
which until that moment required specific parsing algorithms, 
using
a task-independent model (without any further adaptation) that, given an input sequence of words, predicts a sequence of tokens that represent a linearization of a parse tree.

However, while recent efforts
on sequence-to-sequence constituent parsing provided promising results, this trend did not reach state-of-the-art results as it did on other natural language processing tasks such as machine translation \citep{liu-etal-2020-multilingual-denoising,seq2seqMTAAAI} or speech recognition \citep{abs-1904-11660,2020arXiv200605474W}.  In fact, they lagged behind classic parsers based on explicit tree-structured algorithms and supported by a more extensive research background. The gap between 
task-specific constituent parsers and sequence-to-sequence models
cannot be only quantified in terms of accuracy and speed \citep{fernandez-gonzalez-gomez-rodriguez-2020-enriched}, but also in coverage: to the best of our knowledge, 
the latter
have not been applied to discontinuous constituent parsing to date. \added{While we can find numerous studies where transition-based algorithms \citep{coavoux2017,coavoux2019b} and chart-based methods \citep{stanojevic-steedman-2020-span,Corro2020SpanbasedDC} were successfully designed for producing discontinuous structures, there has been no attempt to address the discontinuous constituent parsing task with a sequence-to-sequence neural architecture.}

\added{In order to improve the coverage of sequence-to-sequence constituent parsers, it is necessary to design novel linearization techniques capable of fully encoding discontinuous constituent trees into a sequence of tokens. Taking as starting point previous sequence-to-sequence approaches for continuous constituent parsing  \citep{fernandez-gonzalez-gomez-rodriguez-2020-enriched}, we develop several linearization strategies inspired on how transition-based parsers handle discontinuities. Then, we implement a powerful neural architecture based on the cutting-edge sequence-to-sequence model proposed by \citet{fernandez-astudillo-etal-2020-transition} for graph parsing. The resulting system is not only the first sequence-to-sequence model for discontinuous constituent parsing, but also an accurate approach that delivers a high performance on the main benchmarks.}

\added{Therefore, our main contributions are:}
\begin{itemize}
    \item \textit{The implementation of a novel sequence-to-sequence constituent parser},\footnote{Source code available at \url{https://github.com/danifg/Disco-Seq2seq-Parser}.}
    building
    on the work developed by \citet{fernandez-gonzalez-gomez-rodriguez-2020-enriched} and \citet{fernandez-astudillo-etal-2020-transition}. While the former defines linearizations for continuous parsing that outperform those previously proposed, the latter introduces a deterministic attention technique over a powerful Transformer sequence-to-sequence architecture \citep{fairseq} that significantly increases prediction accuracy. The resulting system outperforms all existing sequence-to-sequence models and is on par with state-of-the-art task-specific constituent parsers.
    \item \textit{Novel linearizations to model discontinuous structures.} We design different strategies to linearize discontinuous constituent trees and test them with the proposed neural architecture, becoming the first sequence-to-sequence model that, not only can produce discontinuous representations, but achieves a competitive performance on the main discontinuous benchmarks: the discontinuous version of the English Penn Treebank (DPTB) \citep{evang-kallmeyer-2011-plcfrs}, and the German NEGRA \citep{Skut1997} and TIGER \citep{brants02} treebanks. 
\end{itemize}

The remainder of this article is organized as follows:
Section~\ref{sec:related} firstly presents previous research work that contributed to model and improve constituent parsing as a sequence-to-sequence problem and, secondly, introduces task-specific constituent parsers proposed so far for dealing with discontinuous trees. Section~\ref{sec:preliminaries} explains how the parsing problem can be cast as a sequence prediction task and other relevant concepts. In Section~\ref{sec:approach}, we detail our approach: we present the novel linearizations developed for generating discontinuities 
and describe 
the proposed neural architecture. In Section~\ref{sec:experiments}, we extensively evaluate our sequence-to-sequence model on continuous and discontinuous treebanks and include a thorough analysis of their performance.  
Lastly, Section~\ref{sec:conclusion} contains a final discussion.

\section{Related work}
\label{sec:related}
\subsection{Constituent parsing as a sequence-to-sequence task}
Since \citet{Vinyals2015} presented the first sequence-to-sequence model for constituent parsing, several 
variants have been proposed
seeking improvements in its performance. These efforts
have
mainly focused on improving either the proposed attentional sequence-to-sequence neural network \citep{Bahdanau2014}, or the linearization technique necessary for casting constituent parsing as a sequence prediction problem, or both.

With respect to the original architecture by \citep{Bahdanau2014} based on recurrent neural networks, several attempts 
have
modified the original design by introducing \textit{deterministic attention} strategies \citep{Kamigaito2017,Ma2017,LiuS2S17,Liu2018,fernandez-gonzalez-gomez-rodriguez-2020-enriched}. These aim to improve the probabilistic attention mechanism (implemented in sequence-to-sequence models to select relevant context) for two purposes: (1) to obtain accuracy gains by deterministically focusing on those input tokens that are crucial for the parsing task and (2) to speed up the decoding process by avoiding 
the need to go over
the whole input sequence when the attentional probabilities are computed. Lastly, \cite{Vaswani2017} propose a novel sequence-to-sequence architecture based on Transformers, which manages to improve both accuracy and speed.

Regarding the linearization strategy, \citet{Vinyals2015} opted for encoding the parse tree from top to bottom by grouping constituents and words by means of brackets. While 
the overwhelming majority of subsequent work
assumed this linearization, there are some exceptions that designed alternative representations. In particular, \citet{Ma2017} and \citet{LiuS2S17} were the first in designing a linearization based on transition-based actions as sequence tokens. The former used actions from the bottom-up transition-based algorithm by \citet{cross-huang-2016-span} to encode constituent trees and proved that it underperforms the original top-down linearization when they are tested under the same conditions; and the latter based the linearization on the top-down transition system of \citet{Dyer2016} and showed that (combined with a specific deterministic attention strategy) this yields some accuracy gains. However, a follow-up study by \citet{fernandez-gonzalez-gomez-rodriguez-2020-enriched} contradicted this last conclusion: they demonstrated that the original bracketed encoding can be represented as sequences of actions from \cite{Dyer2016}'s transition system and, under the same neural network, this representation outperforms the top-down linearization defined by \citet{LiuS2S17}. Additionally, \citet{fernandez-gonzalez-gomez-rodriguez-2020-enriched} proposed a novel linearization method based on the in-order transition system \citep{Liu2017}, notably outperforming all existing sequence-to-sequence constituent parsers.

Finally, it is worth mentioning that there also exists a recent trend of casting constituent parsing as sequence labeling \citep{gomez-rodriguez-vilares-2018-constituent,vilares-gomez-rodriguez-2020-discontinuous}. While these might be considered a kind of sequence-to-sequence methods (where the input and target sequences are constrained to the same length, as each input word is assigned exactly one label as output), they are not usually framed within sequence-to-sequence constituent parsers since the neural architecture used for sequence labeling is simpler than the setup designed by \citep{Bahdanau2014}, obtaining worse accuracy due to the lack of attention mechanisms\footnote{Please note that recent sequence labeling approaches \citep{Vilares_2020,vilares-gomez-rodriguez-2020-discontinuous} also include variants with attention mechanisms that, while notably increasing their accuracy, significantly penalize their speed.} and a larger label dictionary, but being significantly faster due to its simplicity.

\subsection{Discontinuous Constituent Parsing}
Discontinuous structures were initially derived by complex and computationally-expensive 
\textit{chart} parsers based on \textit{Linear Context-Free Rewriting Systems} (LCFRS) \citep{VijWeiJos87} or \textit{Multiple Context Free Grammmars} (MCFGs) \citep{SEKI1991}, which use the CYK algorithm for exact decoding \citep{evang-kallmeyer-2011-plcfrs,Cranenburgh2016,gebhardt-2020-advances}. 
However, the computational complexity of these approaches makes them impractical for long sentences, to the point that all the mentioned parsers evaluate with a cap on sentence length (typically, 40) due to the infeasibility of processing longer sentences.
For this reason, we can also find several variations of these original grammar-based parsers that attempt to reduce their computational cost and make then runnable on long sentences.  For instance, \cite{stanojevic-steedman-2020-span} and \cite{Corro2020SpanbasedDC} speed up decoding by not explicitly defining a set of rules and using a span-based scoring algorithm \citep{stern-etal-2017-minimal}. Additionally, \cite{morbitz2020supertaggingbased} present the first suppertagging-based parser for LCFRS that notably reduces parsing time.

Alternatively, bottom-up
\textit{transition-based} (or \textit{shift-reduce}) parsers, originally restricted to continuous structures \citep{sagae05,zhu-etal-2013-fast}, were extended to generate discontinuities. In particular, some of these parsers incorporate new actions for changing the original token order (allowing to treat discontinuous structures as continuous ones) \citep{maier-2015-discontinuous,maier2016,stanojevic-alhama-2017-neural} or directly processing non-adjacent words  \citep{coavoux2017}; and others opted for designing novel data structures to facilitate building constituents on non-local items \citep{coavoux2019b}.

Finally, several efforts focused on reducing discontinuous constituent parsing into a simpler task. For instance, discontinuous phase-structure trees can be encoded as non-projective dependency trees  and, then, produced by a dependency parser \citep{hall-nivre-2008-dependency,fernandez-gonzalez-martins-2015-parsing}; or represented as a sequence of tags and derived by any tagger \citep{vilares-gomez-rodriguez-2020-discontinuous}. Recently, \cite{fernandezgonzalez2021reducing} also reduced discontinuous into continuous constituent parsing by just accurately reordering input words.

Our work is framed within this last category and the closest approach is the sequence labeling strategy introduced by \cite{vilares-gomez-rodriguez-2020-discontinuous}. However, as mentioned above, they do not use a sequence-to-sequence model to perform the tagging and, since the target sequence shares the same length as the input sentence, they just employ a regular tagger. This constraint requires a complex encoding scheme that results in a remarkably large output dictionary. In contrast, a sequence-to-sequence approach can deal with longer target sequences and, therefore, maintain a smaller dictionary size, outperforming a regular sequence tagger by a wide margin.

\section{Preliminaries}
\label{sec:preliminaries}
\subsection{Continuous linearizations}
\label{sec:contlin}
In order to properly define the sequence prediction problem of \textit{translating} an input sentence of $n$ words $\mathbf{w}=w_1, \dots, w_n$ into a constituent tree $C$,
the latter needs to be encoded (or linearized) as a sequence of tokens $\mathbf{y}=y_1, \dots, y_m$, with $n < m$ in the particular case of sequence-to-sequence constituent parsing.\footnote{As mentioned before, \cite{gomez-rodriguez-vilares-2018-constituent} developed an encoding that, thanks to a large dictionary,  is able to linearize a sentence of $n$ words into a sequence of $n$ tokens.} This conversion must be invertible so that the original tree can be recovered from the sequence of tokens. It is also worth mentioning that, unlike in machine translation, this is an unbalanced sequence prediction task, since target sequences are significantly longer than inputs.

A constituent tree $C$ is composed of words $w_1, \dots w_n$ as leaf nodes and, above them, a number of hierarchically-organized constituents. Each constituent (or phrase) can be represented as a tuple $(X, \mathcal{Y})$ where $X$ is the non-terminal label and $\mathcal{Y}$ is the set of word positions that constitute its yield or span.
Moreover, we define a constituent tree $C$ as \textit{continuous} if the span of every constituent of $C$ is a continuous substring (or, equivalently, a sequence of consecutive word positions). The phase structure tree in Figure~\ref{fig:trees}(a) is an example of a continuous structure. Conversely, if there are at least one constituent of $C$ with a yield composed of non-successive word positions, it will be classified as \textit{discontinuous}. The constituent tree in Figure~\ref{fig:trees}(b) is discontinuous since the span of the constituent (VP, \{0, 2, 3, 4, 6, 7, 8\}) is interrupted by words $\textit{wird}_1$ and $\textit{Wasser}_5$ from constituent (S, \{0, 1, 2, 3, 4, 5, 6, 7, 8\}), resulting in a yield with a sequence of non-consecutive word positions. 

\begin{figure}
\begin{center}
\includegraphics[width=0.9\columnwidth]{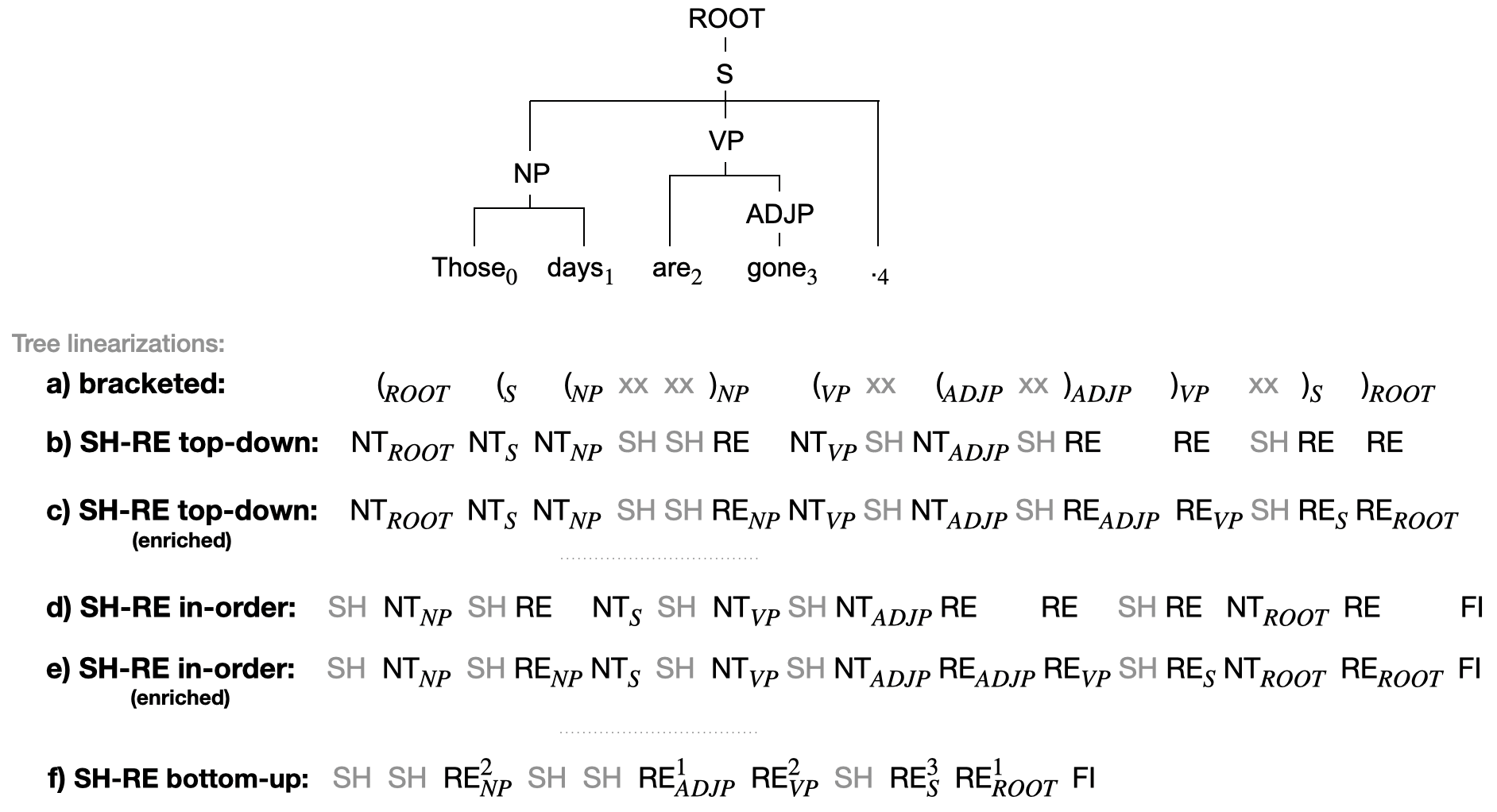}
\end{center}
\caption{Bracketed and shift-reduce (SH-RE) tree linearizations for encoding the continuous constituent tree in Figure~\ref{fig:trees}(a). SH = \textsc{Shift}, NT$_{X}$ = \textsc{Non-Terminal-X}, RE = \textsc{Reduce}, RE$_{X}$ = \textsc{Reduce-X}, RE$^{k}_{X}$ = \textsc{Reduce\#k-X} and FI = \textsc{Finish}.}
\label{fig:linearizations}
\end{figure}

Initially, \citet{Vinyals2015} introduced a simple top-down bracketed encoding for linearizing continuous constituent trees. This consists of defining phrase spans with 
 opening and closing brackets and parametrizing these brackets with non-terminal labels to fully encode constituent information. Additionally, words were normalized by replacing them with a tag \texttt{XX}. An example of this tree linearization is depicted in Figure~\ref{fig:linearizations}(a).

As an alternative, we can find linearizations that use transition-based actions as output tokens (instead of brackets and \texttt{XX}-tags) and represent a syntactic structure as a sequence of transitions \citep{Ma2017,LiuS2S17,fernandez-gonzalez-gomez-rodriguez-2020-enriched}. There exist different shift-reduce transition systems (mainly focused on transition-based parsing) that can also be applied to encode a constituent tree in different manners: from top to bottom, based on an in-order traversal or following a bottom-up strategy. The most successful shift-reduce linearizations are mainly based on two transition systems: the \textit{top-down} transition system \citep{Dyer2016} and the \textit{in-order} shift-reduce algorithm \citep{Liu2017}. Additionally, in this research work, we will also apply and test a linearization based on the bottom-up transition system defined by \cite{nonbinary}. Unlike classic bottom-up approaches \citep{sagae05,zhu-etal-2013-fast}, this transition system does not require to previously binarize the constituent tree and, although it was not used to date as a linearization by any previous work and it was already shown that shift-reduce bottom-up linearizations are not the best option for sequence-to-sequence constituent parsing \citep{Ma2017}, we will also include it for 
comparison purposes.

Transition systems are state machines that traverse a sequence of configurations by means of a set of transitions, until they reach a terminal configuration from which the output tree can be recovered. Thus,
to understand how these specific transition systems work, we need to formally define their \textit{parser configurations} and available \textit{transitions}. 
The top-down algorithm has 
parser configurations of the form {$c=\langle {\Sigma} , {B} \rangle$}, where $B$ is the \textit{buffer} that initially contains all the input words and $\Sigma$ is a \textit{stack}, which is empty in the initial configuration and will store constituents and unprocessed words during the parsing process. Additionally, \new{while top-down terminal configurations are those with an empty buffer and a single element on the stack, the in-order and bottom-up approaches need a third component in their parser configurations for marking whether a parser configuration is terminal or not. This is implemented by a boolean variable $f$ (which will be $\textit{false}$ in the initial configuration) and their configurations will be of the form {$c=\langle {\Sigma} , {B}, f \rangle$}.}


Knowing the configurations, we can now define the transitions for each parser. Firstly, the top-down algorithm provides three transitions (defined in Figure \ref{fig:transitions}) that modify the stack and the buffer to generate a valid constituent tree. Concretely: 
\begin{itemize}
    \item a \textsc{Shift} transition pushes words from the buffer to the stack,
    \item a \textsc{Non-Terminal-X} action adds a non-terminal node \texttt{X} on the stack,
    \item and a \textsc{Reduce} transition pops items from the stack until a non-terminal node is reached and groups all these items as a new constituent on the top of the stack.
\end{itemize}

\begin{figure}
\begin{tabbing}
\hspace{5.2cm}\=\hspace{1cm}\= \kill
\> \textsc{Shift}: 
\> \ \ \ \ \ \ \ $\langle {\Sigma}, {w_i | B}  \rangle
\Rightarrow \langle {\Sigma | w_i} , {B}  \rangle$\\[2mm]
\> \textsc{NT-X}:
\>  \ \ \ \ \ \ \ $\langle {\Sigma}, B \rangle
\Rightarrow \langle {\Sigma | X }, B  \rangle$ { }\\[2mm]
\> \textsc{Reduce}:
\>   \ \ \ \ \ \ \   $\langle {\Sigma} | X| s_k| \dots | s_0 , B \rangle$ $\Rightarrow$ $\langle {\Sigma} | X_{s_k \dots  s_0} , B \rangle$
\end{tabbing}
\caption{Transitions of the top-down transition system (\textsc{NT-X} = \textsc{Non-Terminal-X}).}
\label{fig:transitions}
\end{figure}

Secondly, the in-order algorithm (described in Figure \ref{fig:transitions2}) uses practically the same actions as the top-down variant, but they are applied in a different order and some of them have a different behavior:
\begin{itemize}
    \item a \textsc{Shift} transition is used to move words from the buffer to the stack,
    \item a \textsc{Non-Terminal-X} transition is applied to push a non-terminal node \texttt{X} into the stack, but, unlike in the top-down transition system, it should only be used if the first child node of the future constituent is on top of the stack,
    \item a \textsc{Reduce} transition is available to pop all items from the stack until the first non-terminal node is reached, which is also popped together with the preceding item to build a new constituent on top of the stack,
    \item and, finally, a \textsc{Finish} transition is used to terminate the parsing process.
\end{itemize}

\begin{figure}
\begin{tabbing}
\hspace{4.2cm}\=\hspace{1cm}\= \kill
\> \textsc{Shift}: 
\> \ \ \ \ \ \ \ $\langle {\Sigma}, {w_i | B} , false \rangle
\Rightarrow \langle {\Sigma | w_i} , {B}, false  \rangle$\\[2mm]
\> \textsc{NT-X}:
\>  \ \ \ \ \ \ \ $\langle {\Sigma}, B, false \rangle
\Rightarrow \langle {\Sigma | s_0 | X }, B, false  \rangle$ { }\\[2mm]
\> \textsc{Reduce}:
\>   \ \ \ \ \ \ \   $\langle {\Sigma} | s_k| X| s_{k-1}| \dots | s_0 , B, false \rangle$ $\Rightarrow$ $\langle {\Sigma} | X_{s_k \dots  s_0} , B, false \rangle$\\[2mm]
\> \textsc{Finish}:
\>  \ \ \ \ \ \ \ $\langle {\Sigma}, B, false \rangle
\Rightarrow \langle {\Sigma}, B , true \rangle$ { }
\end{tabbing}
\caption{Transitions of the in-order transition system (\textsc{NT-X} = \textsc{Non-Terminal-X}).}
\label{fig:transitions2}
\end{figure}

Lastly, the non-binary bottom-up transition system provides the following actions (described in Figure \ref{fig:transitions3}):
\begin{itemize}
    \item a \textsc{Shift} transition that pushes words from the buffer to the stack,
    \item a \textsc{Reduce\#k-X} action parameterized with an integer \textit{k} and the non-terminal label \texttt{X} that pops \textit{k} items from the stack and builds a new constituent with all of them on the top of the stack,
    \item and, finally, a \textsc{Finish} transition that marks the end of the process.
\end{itemize}

\begin{figure}
\begin{tabbing}
\hspace{4cm}\=\hspace{1.7cm}\= \kill
\> \textsc{Shift}: 
\> \ \ \ \ \ \ \ $\langle {\Sigma}, {w_i | B} , false \rangle
\Rightarrow \langle {\Sigma | w_i} , {B}, false  \rangle$\\[2mm]
\> \textsc{Reduce\#k-X}:
\>   \ \ \ \ \ \ \   $\langle {\Sigma} | s_{k-1}| \dots | s_0 , B, false \rangle$ $\Rightarrow$ $\langle {\Sigma} | X_{s_{k-1} \dots  s_0} , B, false \rangle$\\[2mm]
\> \textsc{Finish}:
\>  \ \ \ \ \ \ \ $\langle {\Sigma}, B, false \rangle
\Rightarrow \langle {\Sigma}, B , true \rangle$ { }
\end{tabbing}
\caption{Transitions of the non-binary bottom-up transition system.}
\label{fig:transitions3}
\end{figure}

Given the described transition systems, a constituent tree $C$ that represents the syntactic information of the input sentence $\mathbf{w}$ can be encoded into a sequence of shift-reduce actions $\mathbf{y}=y_1, \dots, y_m$ by following a top-down, in-order or bottom-up transition system, as exemplified by sequences (b), (d) and (f) in Figure~\ref{fig:linearizations}, respectively. Depending on the shift-reduce strategy used, the resulting target sequence may have a different length, since each transition system utilizes a different number of transitions to build the same phrase structure.

Additionally, \cite{fernandez-gonzalez-gomez-rodriguez-2020-enriched} noticed that, if \textsc{Reduce} transitions are parameterized with the non-terminal label (\textsc{Reduce-X}), the bracketed \citep{Vinyals2015} and shift-reduce top-down \citep{LiuS2S17} linearizations are equivalent and this \textit{enriched} top-down variant improves prediction accuracy (see an example in Figure~\ref{fig:linearizations}(c)). Applying this idea to the in-order strategy, they also proposed an enriched variant and proved that, while enlarging the output dictionary size, it outperformed all existing tree linearizations  tested on the framework designed by \cite{LiuS2S17} for sequence-to-sequence constituent parsing (Figure~\ref{fig:linearizations}(e) exemplifies this tree linearization).

In spite of using the same transition systems, the main reason why sequence-to-sequence models are so far not obtaining comparable results to task-specific transition-based parsers is the lack of structural constraints explicitly provided by the stack and the buffer, which 
can
help the algorithm to handle and hierarchically organize phrases and words during parsing. Sequence-to-sequence constituent parsers are agnostic to any structural information during decoding and exclusively \textit{translate} an input sentence into a sequence of tokens, which (after a post-processing step) will be converted into a tree structure.

\subsection{Sequence-to-sequence Neural Architecture}
\label{sec:transformer}
\cite{Vinyals2015} propose to address constituent parsing using the attentional sequence-to-sequence neural model defined by \cite{Bahdanau2014}. This work introduces an attention mechanism to the original neural architecture defined by \cite{Sutskever2014} for solving sequence-to-sequence problems and applies it to machine translation. The attention mechanism allows the model to focus, at each time step, on the most relevant information from the input in order to accurately predict the output tokens. This is especially important for handling long sequences, since the prediction accuracy deteriorates as the length of the input sequence increases \citep{cho-etal-2014-learning}. 

More in detail, \cite{Bahdanau2014} define an \textit{encoder-decoder} neural architecture where the \textit{encoder} reads tokens from the input sequence (words in constituent parsing and machine translation) represented as a sequence of vectors $\mathbf{x} = x_1, \dots, x_n $ 
(where each $x_i$ can be obtained from pre-trained word embeddings)
and encodes them as a sequence of \textit{encoder hidden states} $\mathbf{h} = h_1, \dots, h_n $. The encoder was initially implemented  
as a recurrent neural network (RNN), in particular a bidirectional LSTM (BiLSTM) \citep{Graves05} that processes the input in both directions.

Then, at each time step $t$, a \textit{decoder} is used for predicting the next output token $y_t$ from the current decoder hidden state $s_t$, which is generated by a function $f$ fed with the context vector $c_t$ 
and the previous decoder hidden state $s_{t-1}$: 
$$s_t=f(s_{t-1},c_t)$$
Function $f$ is usually implemented as a unidirectional LSTM \citep{Hochreiter97} and the context vector $c_t$ 
is computed at time step $t$ as follows:
$$c_t = \sum_{i=1}^{n}\alpha_{ti}h_i, \ \ 
\alpha_{ti} = \frac{\textit{exp}(\beta_{ti})}{\sum_{k=1}^{n} \textit{exp}(\beta_{tk})}, \ \ \beta_{ti} = g(s_{t-1}, h_i)$$
where $g$ is a scoring function (implemented as a feed-forward neural network jointly trained with the other components) that computes scores between each input token $x_i$ (encoded as $h_i$) and the previous decoder hidden state $s_{t-1}$. Then, a probability distribution over the whole input is computed using a softmax function, reflecting in $c_t$ the weight of each input token in the prediction of the current output token $y_t$.

Especially in constituent parsing, it has been observed that by using \textit{deterministic} attention mechanisms the performance of the original sequence-to-sequence model increases \citep{Kamigaito2017,Ma2017,Liu2018,fernandez-gonzalez-gomez-rodriguez-2020-enriched}. The most successful variant was developed by  
\cite{LiuS2S17} and is based on the top-down shift-reduce linearization. They propose to use two separate attention models and, therefore, calculate two different context vectors instead of just $c_t$. 
For this purpose, the model splits the input into two variable-length segments
obtained by dividing the input sequence by index $p$, which in the initial decoding step points to the first input token,
and then is incremented to point to the next word whenever a \textsc{Shift} transition is predicted.
Then, vector $c^l_t$ is computed over the left segment $w_1, \dots, w_p$ and, $c^r_t$, over the right segment $w_{p+1}, \dots, w_n$ as follows:
$$c^l_t = \sum_{i=1}^{p}\alpha_{ti}h_i,\ \ \  c^r_t = \sum_{i=p+1}^{n}\alpha_{ti}h_i$$

\noindent This approach led to notable accuracy gains \citep{fernandez-gonzalez-gomez-rodriguez-2020-enriched}: 
on the one hand,
it
intuitively models a stack ($c^l_t$) and a buffer ($c^r_t$) over the input that are modified when a \textsc{Shift} transition is applied. 
On the other hand, it
provides a deterministic alignment between input words and \textsc{Shift} tokens 
that is
crucial for choosing the most relevant context information at each time step.

In the last few years, these RNN sequence-to-sequence models were 
substituted by 
a Transformer architecture thanks to the remarkable performance presented by \cite{Vaswani2017}. The authors not only prove that Transformers notably outperform RNNs on machine translation, but they also apply this new architecture on constituent parsing, obtaining promising results. Both the encoder and decoder are implemented with Transformers, which provide a multi-head self-attention mechanism that is more powerful than the attention techniques developed in RNN approaches. 

More in detail, this novel architecture starts by injecting a positional encoding to each input word representation $x_i$ necessary for handling sequences, as, unlike RNNs, Transformers contain no recurrence and do not have any built-in notion of 
sequential order.
After that, the encoder implemented by six Transformers generates the sequence of encoder hidden states $\mathbf{h} = h_1, \dots, h_n $ for the input sequence. To do this, each Transformer implements a \textit{multi-head self-attention layer} that is 
composed of
several parallel \textit{attention heads}, which will score the relevance of a specific word with respect to the other words in the sentence. In particular, each head 
computes attention vectors $z_{i}$ for each input word representation $x_i$ as a weighted sum of linearly transformed input vectors:

$$z_i = \sum_{j=1}^{n}\alpha_{ij}(x_jW^V), \ \ 
\alpha_{ij} = \frac{\textit{exp}(\beta_{ij})}{\sum_{k=1}^{n} \textit{exp}(\beta_{ik})}, \ \ \beta_{ij} = \frac{(x_iW^Q)(x_j{W^K}) ^T}{\sqrt{d}}$$

\noindent where $W^Q$, $W^K$ and $W^V$ are parameter matrices unique per attention head,
 $d$ is the dimension of the resulting vector $z_i$ and $\beta_{ij}$ is computed by a compatibility function (implemented as an efficient scaled dot product) that compares two input words. The multi-head self-attention layer is followed by a feed-forward network for finally generating encoder hidden states $\mathbf{h}$. Unlike RNNs, this process can be easily parallelized, speeding up Transformers performance.

\begin{figure*}
\centering
\includegraphics[width=0.6\textwidth]{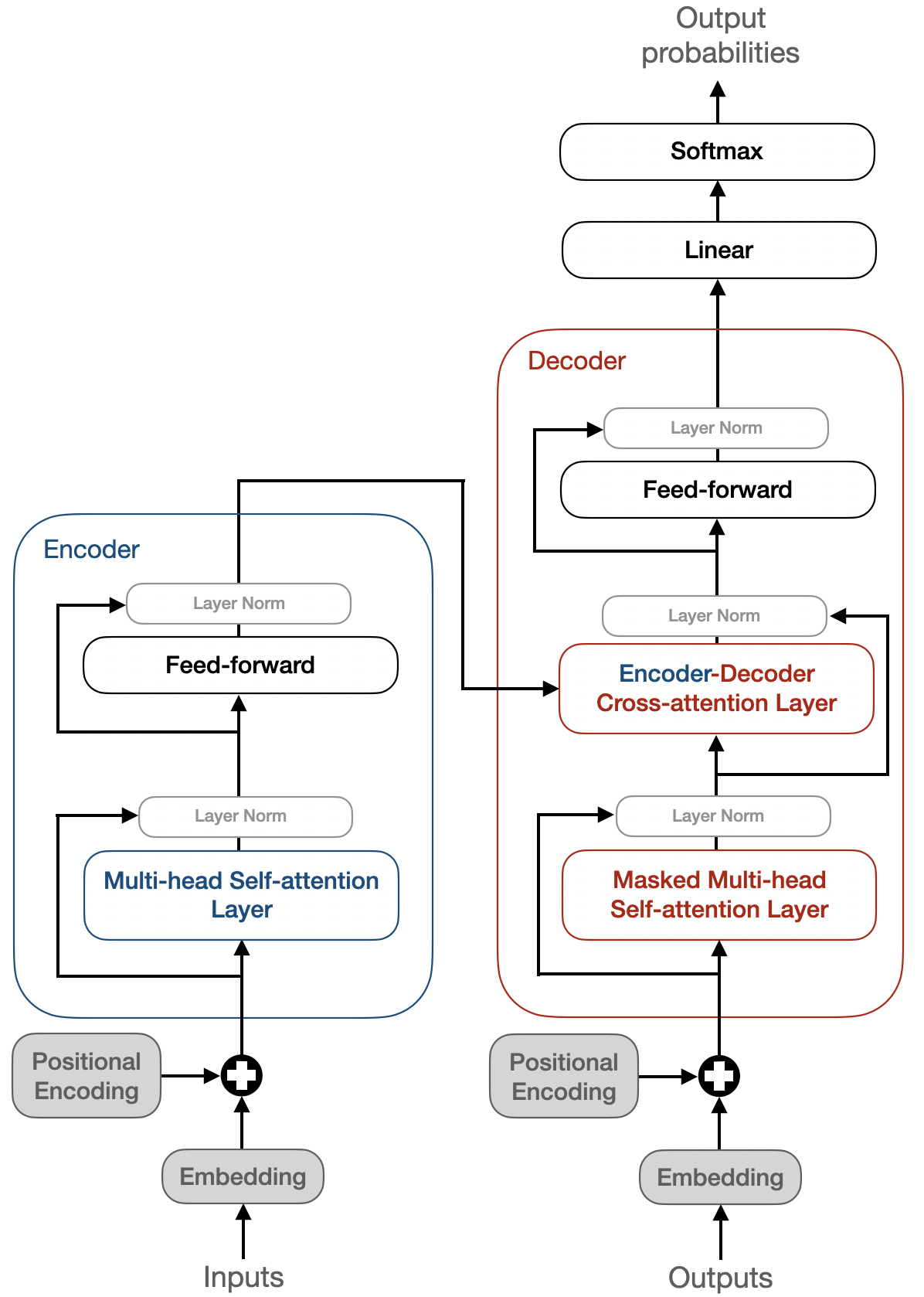}
\caption{\added{Neural architecture for the Transformer sequence-to-sequence model introduced by  \cite{Vaswani2017}. Please note that \textit{Layer Norm} stands for the layer normalization technique introduced by \cite{layernormalization}.}}
\label{fig:network}
\end{figure*}

Regarding the decoder, it is also implemented by six Transformers, but each one has an additional component. Apart from a \textit{masked multi-head self-attention mechanism} that works practically the same as the encoder (which is used to encode previously-generated output tokens $\mathbf{y} = y_1, \dots, y_{t-1}$ into a sequence $\mathbf{q} = q_0, \dots, q_{t-1}$ at time step $t$), it implements a posterior \textit{encoder-decoder cross-attention layer} that computes the compatibility between each target token with each input word. More in detail, this cross-attention module is also composed of
several attention heads that, given the sequences of encoder and decoder hidden states $\mathbf{h}$ and $\mathbf{q}$, generate at each time step $t$ an attention vector $z_t$ as follows: 
$$z_t = \sum_{i=1}^{n}\alpha_{ti}(h_iW_d^V), \ \ 
\alpha_{ti} = \frac{exp(\beta_{ti})}{\sum_{k=1}^{n} exp(\beta_{tk})}, \ \ \beta_{ti} = \frac{(q_{t-1}W_d^Q)(h_i{W_d^K}) ^T}{\sqrt{d}}$$

\noindent where $W_d^Q$, $W_d^K$ and $W_d^V$ are parameter matrices, $d$ is the dimension of the resulting vector $z_t$ and $\beta_{ti}$ computes the interaction between the last predicted token encoding ($q_{t-1}$, which represents target token history) with each word from the input sequence (represented by its encoder hidden state $h_i$). The attention vectors $z_t$ computed by each head will be combined and used by posterior linear and softmax layers to finally generate the output token $y_t$. \added{Please see in Figure~\ref{fig:network} a sketch of the described neural architecture.}

Recently, \citet{fernandez-astudillo-etal-2020-transition} implemented the idea introduced by \cite{LiuS2S17} into this Transformer sequence-to-sequence architecture for dependency and AMR (Abstract Meaning Representation) parsing. \added{To that end, they followed the current trend of modifying \cite{Vaswani2017}'s architecture by implementing dedicated attention heads to focus on one or several tokens from the input and, thus, encode local relevant information \citep{strubell-etal-2018-linguistically,xu-etal-2019-leveraging}. In particular,} instead of using an index $p$ to delimitate the stack and buffer, they propose to specialize two of the attention heads (from the Transformer decoder's cross-attention layer): one 
for just attending input words that should be into the stack according to the shift-reduce tokens applied so far, and the other for exclusively considering words that are left in the buffer.
With these two dedicated heads for the decoding, they manage to substantially improve over the original setup by \cite{Vaswani2017} on these two graph parsing tasks. In this article, we accordingly modify this 
recent
sequence-to-sequence model 
for handling
 constituent parsing and accurately produce continuous and discontinuous phrase structure trees.

\section{Discontinuous Sequence-to-sequence Parsing}
\label{sec:approach}
\subsection{Discontinuous Linearizations}
\label{sec:linearizations}
Since all transition systems (and, therefore, the resulting tree linearizations) explained in Section~\ref{sec:contlin} are restricted to continuous structures, 
we need to extend them
to handle discontinuities. For that purpose, novel transition systems must be defined so that they can be used for linearizing discontinuous structures and extending the coverage of sequence-to-sequence models  (currently constrained to continuous structures) to any kind of constituent trees.

Please note that, while it can be argued that there already exist transition systems that can produce discontinuous constituent trees \citep{maier-2015-discontinuous,maier2016,stanojevic-alhama-2017-neural,coavoux2017}, all of these follow a binary bottom-up strategy that is not the most adequate for sequence-to-sequence constituent parsing as shown by \cite{Ma2017}. However, for completeness of comparison, we also include in our experiments a discontinuous extension of the non-binary bottom-up transition system by \cite{nonbinary}, which was shown to be superior to the binary variants in continuous transition-based parsing.

\begin{figure}
\begin{tabbing}
\hspace{2.7cm}\=\hspace{1.7cm}\= \kill
\> a) \textsc{Swap}: 
\> \ \ \ \ \ \ \ $\langle {\Sigma | w_1 | w_0}, {B} , false \rangle
\Rightarrow \langle {\Sigma | w_0} , {w_1 | B}, false  \rangle$\\[2mm]
\> b) \textsc{Swap\#k}:
\>   \ \ \ \ \ \ \   $\langle {\Sigma} | w_{k}| \dots | w_1 | w_0 , B, false \rangle$ $\Rightarrow$ $\langle {\Sigma | w_0} , {w_k | \dots | w_1 | B}, false \rangle$\\[2mm]
\> c) \textsc{Shift\#k}:
\> \ \ \ \ \ \ \ $\langle {\Sigma}, {w_0 | \dots | w_{n-1}} , false \rangle
\Rightarrow \langle {\Sigma | w_k} , {w_0 | \dots |w_{k-1}| w_{k+1} | \dots | w_{n-1}}, false  \rangle$
\end{tabbing}
\caption{Available transitions for reordering words from the input sentence.}
\label{fig:transitions4}
\end{figure}

Firstly, we define new discontinuous transition systems by adding to the top-down \citep{Dyer2016}, in-order \citep{Liu2017} and the mentioned non-binary bottom-up algorithms a \textsc{Swap} transition. This action (initially proposed for transition-based constituent parsing by \cite{versley-2014-experiments})\footnote{The concept of a \textsc{Swap} transition was initially introduced for non-projective dependency parsing by \cite{nivre-2009-non} and then adapted by \cite{versley-2014-experiments} to constituent parsing.} is used to reorder the original sentence by moving the second word on top of the stack back to 
the buffer (as detailed in Figure~\ref{fig:transitions4}(a)). Thanks to this online reordering of the input during parsing, any discontinuous structure can be created with the available continuous transitions. This relies on the fact that any discontinuous constituent tree can be transformed into a continuous variant by just changing the order of tokens. \added{For instance, the discontinuous tree in Figure~\ref{fig:trees}(b) can be converted into a continuous one by moving the word \textit{wird}$_1$ before the word \textit{Allerdings}$_0$, and the word \textit{Wasser}$_5$ after the word \textit{verteilt}$_8$.}
We show in Figure~\ref{fig:example} how the in-order transition system extended with the \textsc{Swap} transition is able to handle the discontinuities of the tree in Figure~\ref{fig:trees}(b) by means of buffer and stack structures.

\begin{figure*}
\begin{center}
\begin{tabular}{@{\hskip 0.1pt}l@{\hskip 0.1pt}c@{\hskip 0.1pt}c@{\hskip 0.1pt}}
\toprule
Transition & Stack & Buffer \\
\midrule
\vspace*{3pt}
 & [ ] & [ Allerdings$_0$, wird$_1$, ... , \textbf{.} ]   \\
\vspace*{3pt}
\textsc{Shift} & [ Allerdings$_0$ ] & [ wird$_1$, in$_2$, ... , \textbf{.} ]   \\
\vspace*{3pt}
\textsc{Non-Terminal-VP} & [ Allerdings$_0$, \textsc{VP} ] & [ wird$_1$, in$_2$, ... , \textbf{.} ] \\
\vspace*{3pt}
\textsc{Shift} & [ Allerdings$_0$, \textsc{VP}, wird$_1$ ] & [ in$_2$, bestimmten$_3$, ... , \textbf{.} ] \\
\vspace*{3pt}
\textsc{Shift} & [ Allerdings$_0$, \textsc{VP}, wird$_1$, in$_2$ ] & [ bestimmten$_3$, Vierteln$_4$, ... , \textbf{.} ] \\
\vspace*{3pt}
\textsc{Swap} & [ Allerdings$_0$, \textsc{VP}, in$_2$ ] & [ wird$_1$, bestimmten$_3$, Vierteln$_4$, ... , \textbf{.} ] \\
\vspace*{3pt}
\textsc{Non-Terminal-PP} & [ Allerdings$_0$, \textsc{VP}, in$_2$, \textsc{PP} ] & [ wird$_1$, bestimmten$_3$, Vierteln$_4$, ... , \textbf{.} ] \\
\vspace*{3pt}
\textsc{Shift} & [ Allerdings$_0$, \textsc{VP}, in$_2$, \textsc{PP}, wird$_1$ ] & [ bestimmten$_3$, Vierteln$_4$, ... , \textbf{.} ] \\
\vspace*{3pt}
\textsc{Shift} & [ Allerdings$_0$, \textsc{VP}, in$_2$, \textsc{PP}, wird$_1$, bestimmten$_3$ ] & [ Vierteln$_4$, Wasser$_5$, ... , \textbf{.} ] \\
\vspace*{3pt}
\textsc{Swap} & [ Allerdings$_0$, \textsc{VP}, in$_2$, \textsc{PP}, bestimmten$_3$ ] & [ wird$_1$, Vierteln$_4$, Wasser$_5$, ... , \textbf{.} ] \\
\vspace*{3pt}
\textsc{Shift} & [ Allerdings$_0$, \textsc{VP}, in$_2$, \textsc{PP}, bestimmten$_3$, wird$_1$ ] & [ Vierteln$_4$, Wasser$_5$, ... , \textbf{.} ] \\
\vspace*{3pt}
\textsc{Shift} & [ Allerdings$_0$, \textsc{VP}, in$_2$, \textsc{PP}, bestimmten$_3$, wird$_1$, Vierteln$_4$ ] & [ Wasser$_5$, aus$_6$, ... , \textbf{.} ] \\
\vspace*{3pt}
\textsc{Swap} & [ Allerdings$_0$, \textsc{VP}, in$_2$, \textsc{PP}, bestimmten$_3$, Vierteln$_4$ ] & [ wird$_1$, Wasser$_5$, ... , \textbf{.} ] \\
\vspace*{3pt}
\textsc{Reduce} & [ Allerdings$_0$, \textsc{VP}, \textsc{PP}$_{(in_2 \ bestimmten_3 \ Vierteln_4)}$ ] & [ wird$_1$, Wasser$_5$, ... , \textbf{.} ] \\
\vspace*{3pt}
\dots & \dots   &  \dots  \\
\bottomrule
\end{tabular}
\caption{Transition sequence 
\new{for partially producing the
discontinuous tree in Figure~\ref{fig:trees}(b)} using
the in-order + \textsc{Swap} transition system in transition-based constituent parsing.} \label{fig:example}   
\end{center}
\end{figure*}

The main drawback of adding the \textsc{Swap} action to continuous transition systems is that it tends to produce considerably long transition sequences 
and, when used as a linearization technique for discontinuous constituent trees, it will generate such long target sequences that it might harm accuracy of prediction. To address this, we also develop two variants based on two transitions already studied in shift-reduce parsing:

\begin{itemize}
    \item a \textsc{Swap\#k} transition \citep{maier-2015-discontinuous} (detailed in Figure~\ref{fig:transitions4}(b)), which is equivalent to applying $k$ \textsc{Swap}s in a row, with \textsc{Swap\#1} being equivalent to applying a single \textsc{Swap} action.
    \item a \textsc{Shift\#k} action \citep{maier2016} (described in Figure~\ref{fig:transitions4}(c)), which moves the $k$th word in the buffer to the stack, with \textsc{Shift\#0} being equivalent to the regular \textsc{Shift} action. In Figure~\ref{tab:example2}, we present an example of how the \textsc{Shift\#k} transition works on the in-order transition system in a shift-reduce parsing process with a buffer and a stack.
\end{itemize}

\noindent While adding the \textsc{Swap\#k} transition will have a minor impact in shortening the resulting sequences, extending the original in-order transition system with the \textsc{Shift\#k} action will lead to a transition sequence 
with the same number of items as if the encoded tree were continuous.

\begin{figure*}
\begin{center}
\begin{tabular}{@{\hskip 0.1pt}l@{\hskip 0.1pt}c@{\hskip 0.1pt}c@{\hskip 0.1pt}}
\toprule
Transition & Stack & Buffer \\
\midrule
\vspace*{3pt}
 & [ ] & [ Allerdings$_0$, wird$_1$, ... , \textbf{.} ]   \\
\vspace*{3pt}
\textsc{Shift\#0} & [ Allerdings$_0$ ] & [ wird$_1$, in$_2$, ... , \textbf{.} ]   \\
\vspace*{3pt}
\textsc{Non-Terminal-VP} & [ Allerdings$_0$, \textsc{VP} ] & [ wird$_1$, in$_2$, ... , \textbf{.} ] \\
\vspace*{3pt}
\textsc{Shift\#1} & [ Allerdings$_0$, \textsc{VP}, in$_2$ ] & [ wird$_1$, bestimmten$_3$, Vierteln$_4$, ... , \textbf{.} ] \\
\vspace*{3pt}
\textsc{Non-Terminal-PP} & [ Allerdings$_0$, \textsc{VP}, in$_2$, \textsc{PP} ] & [ wird$_1$, bestimmten$_3$, Vierteln$_4$, ... , \textbf{.} ] \\
\vspace*{3pt}
\textsc{Shift\#1} & [ Allerdings$_0$, \textsc{VP}, in$_2$, \textsc{PP}, bestimmten$_3$ ] & [ wird$_1$, Vierteln$_4$, Wasser$_5$, ... , \textbf{.} ] \\
\vspace*{3pt}
\textsc{Shift\#1} & [ Allerdings$_0$, \textsc{VP}, in$_2$, \textsc{PP}, bestimmten$_3$, Vierteln$_4$ ] & [ wird$_1$, Wasser$_5$, ... , \textbf{.} ] \\
\vspace*{3pt}
\textsc{Reduce} & [ Allerdings$_0$, \textsc{VP}, \textsc{PP}$_{(in_2 \ bestimmten_3 \ Vierteln_4)}$ ] & [ wird$_1$, Wasser$_5$, ... , \textbf{.} ] \\
\vspace*{3pt}
\dots & \dots   &  \dots  \\
\bottomrule
\end{tabular}
\caption{Transition sequence for \new{for partially building the}
discontinuous tree in Figure~\ref{fig:trees}(b) using
the in-order + \textsc{Shift\#k} transition system in transition-based constituent parsing.} \label{tab:example2}   
\end{center}
\end{figure*}

\begin{figure}
\begin{center}
\includegraphics[width=0.9\columnwidth]{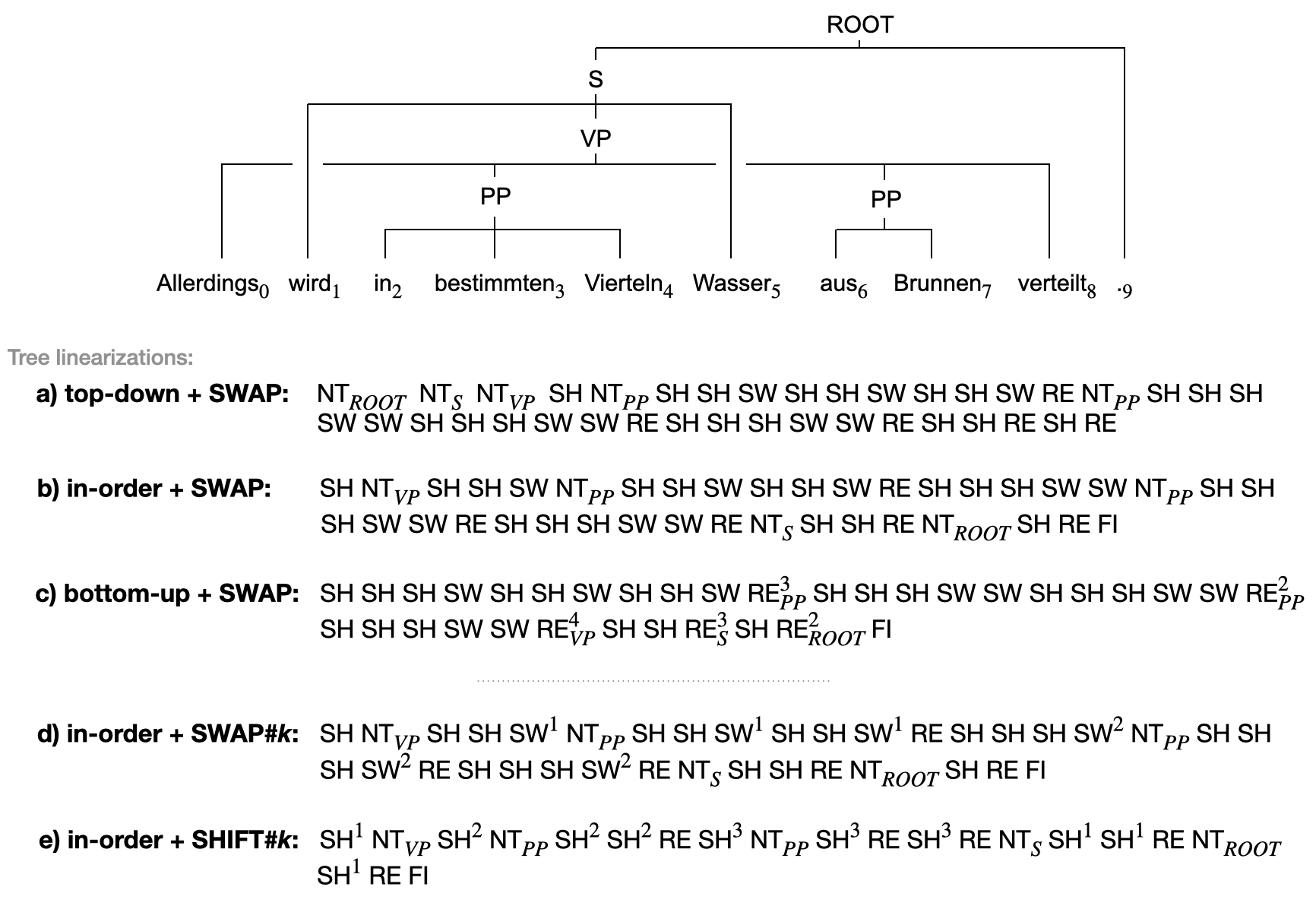}
\end{center}
\caption{Novel shift-reduce linearizations for encoding the discontinuous constituent tree in Figure~\ref{fig:trees}(b). SH = \textsc{Shift}, SH$^k$ = \textsc{Shift\#k}, SW = \textsc{SWAP}, SW$^k$ = \textsc{SWAP\#k}, NT$_{X}$ = \textsc{Non-Terminal-X}, RE = \textsc{Reduce}, 
RE$^{k}_{X}$ = \textsc{Reduce\#k-X} and FI = \textsc{Finish}.}
\label{fig:discolinearizations}
\end{figure}

All these novel discontinuous transition systems can be used as linearization techniques for casting discontinuous constituent trees as sequences of tokens (which then can be used for training a sequence-to-sequence model). For instance, in Figures~\ref{fig:discolinearizations}(a), (b) and (c), we can see the resulting linearizations of a discontinuous tree by the top-down, in-order and bottom-up algorithms extended with the \textsc{Swap} transition, respectively. Additionally,  Figures~\ref{fig:discolinearizations}(d) and (e) show the discontinuous variant of the in-order linearization with the \textsc{Swap\#k} and \textsc{Shift\#k} transitions, respectively. \new{Although an analysis in this regard is included in Section~\ref{sec:analysis},} it can be noticed in this example how the presence of \textsc{Swap} tokens notably lengthens tree linearizations in comparison to the variant with the \textsc{Shift\#k} transition, which has the same number of tokens as linearizing a continuous tree. While \textsc{Swap\#k} and \textsc{Shift\#k} transitions can also be applied to the top-down and bottom-up linearization methods, we only test these variants on the best-performing strategy,
which is the one
based on the in-order tree linearization (as will be seen in Section~\ref{sec:experiments}).

We opted for not using the enriched variants proposed by \cite{fernandez-gonzalez-gomez-rodriguez-2020-enriched} (which parametrize \textsc{Reduce} actions in the top-down and in-order transition systems) since, as we will see in Section~\ref{sec:experiments}, the accuracy of the regular versions is on par with these enhanced variants on the proposed neural architecture (while requiring a smaller output dictionary).

\added{Finally, please note that the proposed transition systems are exclusively used for linearizing constituent trees and defining the target sequence of tokens for our sequence-to-sequence model. This contrasts with task-specific transition-based parsers, which leverage data structures (two or more stacks) to explicitly build partial constituent trees at each step of the parsing process. In addition, it is also worth mentioning that 
the presented discontinuous transition-based algorithms were never 
proposed
before 
as far as we know
and, while they might certainly achieve a good performance 
under a traditional
shift-reduce parsing
implementation,
this is out of the scope of this research work and we will exclusively apply them as tree linearization strategies.}

\subsection{Neural Architecture}
\label{sec:arch}
Based on the approach proposed by \citep{fernandez-astudillo-etal-2020-transition} for dependency and AMR parsing, we present a Transformer sequence-to-sequence architecture for unrestricted constituent parsing. Unlike the original work by \cite{Vaswani2017} (where all \textit{heads} of each Transformer decoder's cross-attention layer attend to the whole input), one specialized head is exclusively applied over input words that should be into the stack according to the current time step $t$ (following a shift-reduce parsing process), and another dedicated head will focus only on those words that are still left in the buffer in $t$. 
In that way, while not explicitly using data structures to process the input sentence (as done by regular transition-based parsers), some structural information is \textit{deterministically} induced to the sequence-to-sequence model, 
with the purpose of substantially increasing
parsing performance as shown 
on other syntactic formalisms
by \cite{fernandez-astudillo-etal-2020-transition}.


\added{The implementation proposed by \cite{fernandez-astudillo-etal-2020-transition} was exclusively focused on dependency and AMR parsing, where the buffer and the stack of a purely transition-based dependency parser can only contain words that belong to partial graph structures. In contrast, transition-based constituent parsers process tree structures and, in addition to nodes corresponding to words, they
also have to
push 
non-terminal nodes into the stack. These are necessary for naming constituents and are especially required for the top-down and in-order transition systems (which make use of the \textsc{Non-Terminal-X} transition for that purpose). In addition, \textsc{Reduce} actions in transition-based constituent parsing build partial subtrees, affecting several words of the input sequence (while \textsc{Reduce} transitions in dependency parsing only affect one single word). Therefore, all these specifics must be taken into consideration for representing the behavior of the data structures in constituent parsing.}

\begin{figure}
\begin{center}
\includegraphics[width=0.9\columnwidth]{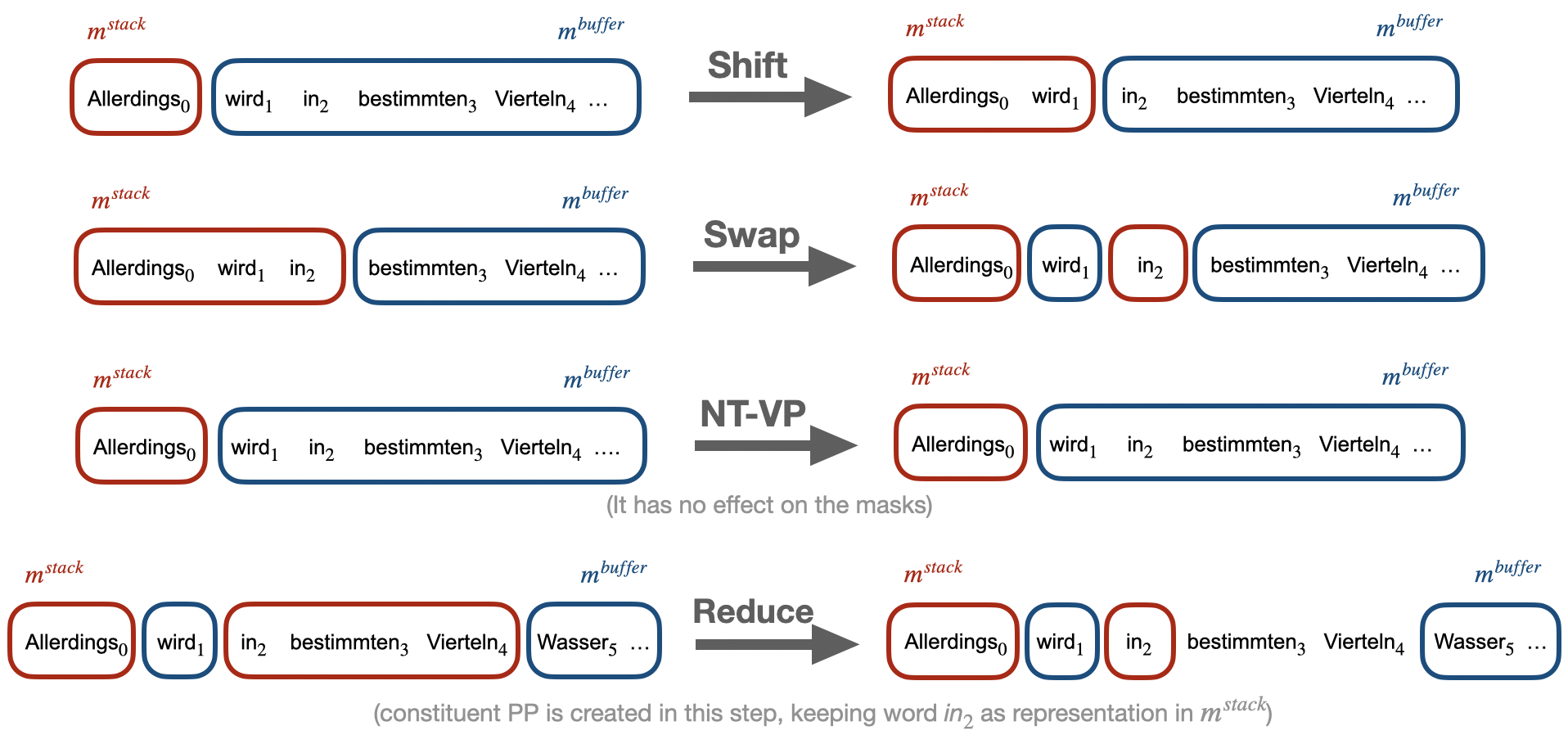}
\end{center}
\caption{Modifications on the $m^{\textit{stack}}$ and $m^{\textit{buffer}}$ masks for modelling the changes produced by some transitions (of the inorder + \textsc{Swap} transition system) on the buffer and stack structures during the parsing process described in Figure~\ref{fig:example}.}
\label{fig:masks}
\end{figure}

More in detail, to implement these dedicated stack and buffer heads of the Transformer decoder's cross-attention mechanism, two masks $m^{\textit{stack}}$ and $m^{\textit{buffer}}$ over the input are defined. Additionally, they must be
updated at each time step $t$ based on the output token predicted in time step $t-1$ and 
accordingly to a shift-reduce \added{constituent parsing} standpoint: 
\begin{itemize}
    \item If a \textsc{Shift} transition was the previous output token, $m^{\textit{buffer}}$ will mask out the first masked word and this will be included in $m^{\textit{stack}}$. The prediction of a \textsc{Shift\#k} token will have a similar behavior, but affecting the word in the $k$th position of $m^{\textit{buffer}}$.
    \item Generating a \textsc{Swap} token in the previous step will modify $m^{\textit{stack}}$ by masking out the second-to-last word, and $m^{\textit{buffer}}$ by adding this word. This modification will be applied $k$ times if a \textsc{Swap\#k} action was the previous output token.
    \item The \textsc{Non-terminal-X} token will have no effect into either $m^{\textit{stack}}$ or $m^{\textit{buffer}}$, since heads can only attend to input words and non-terminal node \texttt{X} is not a token from the input sequence.
    \item Predicting a \textsc{Reduce} token (including \textsc{Reduce-X} and \textsc{Reduce\#k-X}) will mask out all words from $m^{\textit{stack}}$ that form the resulting constituent, except the first token that will be kept in $m^{\textit{stack}}$ as a representation of the reduced constituent.
\end{itemize}
In Figure~\ref{fig:masks}, we graphically depict how these masks change each time a shift-reduce token is predicted and how they model the content of the stack and buffer during the decoding process.
 
Given the 
mask $m^{\textit{stack}}_t$ \new{(implemented as a vector of $-\infty$ or 0 values)}
at time step $t$, the equation proposed by \citep{Vaswani2017} (and introduced in Section~\ref{sec:transformer}) is modified for computing the attention head $z^{stack}_t$ 
that exclusively attends words in the stack following a shift-reduce point of view:

$$z^{stack}_t = \sum_{i=1}^{n}\alpha_{ti}(h_iW_d^V), \ \ 
\alpha_{ti} = \frac{exp(\beta_{ti})}{\sum_{k=1}^{n} exp(\beta_{tk})}, \ \ \beta_{ti} = \frac{(q_{t-1}W_d^Q)(h_i{W_d^K}) ^T}{\sqrt{d}} + m^{\textit{stack}}_{ti}$$

\noindent The same computation is done for, given 
mask $m^{\textit{buffer}}_t$, obtaining $z^{\textit{buffer}}_t$ that attends to input tokens still in the buffer at time step $t$ according to a transition-based parsing process:

$$z^{\textit{buffer}}_t = \sum_{i=1}^{n}\alpha_{ti}(h_iW_d^V), \ \ 
\alpha_{ti} = \frac{exp(\beta_{ti})}{\sum_{k=1}^{n} exp(\beta_{tk})}, \ \ \beta_{ti} = \frac{(q_{t-1}W_d^Q)(h_i{W_d^K}) ^T}{\sqrt{d}} + m^{\textit{buffer}}_{ti}$$

All shift-reduce linearizations described in Sections~\ref{sec:contlin} and \ref{sec:linearizations} can be learnt by this neural architecture without further modifications. Additionally, new shift-reduce linearizations can be included by just adapting the modifications to stack and buffer masks if new transitions are incorporated.

We also want to point out the importance of inducing some structural information with the proposed deterministic attention technique especially in discontinuous parsing, since the lack of explicit structures (crucial for handling the word reordering with the \textsc{Swap} action) may penalize parsing performance. 

Finally, we adopt the same encoder and first component of the decoder (named as \textit{masked multi-head self-attention}) from \citep{Vaswani2017}, just applying the mentioned modifications to two of the heads from the \textit{encoder-decoder cross-attention multihead attention} module.

\section{Experiments}
\label{sec:experiments}
\subsection{Setup}
\paragraph{Data} For properly testing our approach, we include both continuous and discontinuous constituent treebanks. Concretely, the continuous 
English Penn Treebank (PTB) \citep{marcus93} and its discontinuous version (DPTB) \citep{evang-kallmeyer-2011-plcfrs} with standard 
splits defined as follows: Sections 2 to 21 for training, 22 for development and 23 for testing. We also include in the evaluation German treebanks with a higher degree of discontinuity: NEGRA \citep{Skut1997} and TIGER \citep{brants02} with commonly-used splits defined by \citet{dubey2003} and \citet{seddah-etal-2013-overview}, respectively. In all cases, we discard Part-of-Speech (PoS) tag information and the number of samples per treebank split are detailed in Table~\ref{tab:treebanks}.

\begin{table}[h]
\centering
\begin{tabular}{@{\hskip 0pt}lccc@{\hskip 0pt}}
\toprule
\textbf{Treebank} & \textbf{Training} & \textbf{Dev} & \textbf{Test} \\
\midrule
PTB/DPTB & 39,832 & 1,700 & 2,416 \\
NEGRA & 18,602 & 1,000 & 1,000 \\
TIGER & 40,472 & 5,000 & 5,000 \\
\bottomrule
\end{tabular}
\setlength{\abovecaptionskip}{4pt}
\caption{Number of samples per treebank split.}
\label{tab:treebanks}
\end{table}

\paragraph{Implementation} Following \citep{fernandez-astudillo-etal-2020-transition}, the proposed neural architecture was developed based on the neural model by \cite{ott-etal-2018-scaling}. The latter implements the Transformer model \citep{Vaswani2017} in Pytorch on the \texttt{fairseq-py} toolkit\footnote{\url{https://github.com/pytorch/fairseq}} and applies it for sequence-to-sequence machine translation. Since constituent parsing can be cast as a sequence-to-sequence task, their model can be easily adapted to our specific problem with minor modifications. In fact, we do not undertake further parameter optimization to our specific task and directly use hyper-parameters reported by \citet{fernandez-astudillo-etal-2020-transition} for 
cross-entropy training with label smoothing and with the learning rate increasing linearly for 4,000 warm-up 
updates to 5$e^{-4}$ and then being decayed proportionally to the inverse square root of the number of steps. 
These hyperparameters are summarized in
Table~\ref{tab:hyperparameters} 
and we refer the reader to
\citep{ott-etal-2018-scaling} for further 
implementation details. Finally, we use average weights of the three best checkpoints on the development splits as 
final models, and beam 10 for decoding.

\begin{table}[h]
\begin{footnotesize}
\centering
\begin{tabular}{@{\hskip 0pt}lc@{\hskip 0pt}}
\toprule
\multicolumn{2}{c}{\textbf{Architecture and optimizer hyper-parameters}} \\
\midrule
Transformer Encoder layers & 6 \\ 
Transformer Encoder size & 256 \\
Transformer Decoder layers & 6 \\ 
Transformer Decoder size & 256 \\
Heads per self-attention layer & 4 \\ 
RoBERTa embedding dimension & 1024\\
GottBERT embedding dimension & 768\\
Dropout & 0.3 \\
Optimizer & Adam \citep{Adam} \\
Loss & cross-entropy \\
$\beta_1$ & 0.9 \\
$\beta_2$ & 0.98 \\
Learning rate & 5$e^{-4}$ \\
Learning rate scheduler & Inverse square root \\
Warm-up initial learning rate & 1$e^{-7}$ \\
Warm-up updates & 4000 \\
Minimum learning rate & 1$e^{-9}$ \\
Label smoothing & 0.01 \\
Batch size & 3584 \\
Training epochs & 80 \\
\bottomrule
\end{tabular}
\setlength{\abovecaptionskip}{4pt}
\caption{Model hyper-parameters.}
\label{tab:hyperparameters}
\end{footnotesize}
\end{table}

\paragraph{Pre-trained Embeddings} For initializing word embeddings, we use fixed weights extracted from the RoBERTa-large \citep{roberta} and GottBERT-base \citep{gottbert} pre-trained language models for English and German, respectively. We apply average weights from wordpieces when required and do not fine-tune word embeddings during training.

\paragraph{Evaluation} We follow standard practice for evaluation and report F-scores with the EVALB script\footnote{\url{https://nlp.cs.nyu.edu/evalb/}} for the continuous PTB  (discarding punctuation), and DISCODOP\footnote{\url{https://github.com/andreasvc/disco-dop}} 
\citep{Cranenburgh2016} for discontinuous treebanks (ignoring punctuation and root symbols). The latter also delivers a Discontinuous F-score (DF1) measured only on discontinuous constituents. For each experiment, we report the average score and standard deviation over three executions with different seeds.

\paragraph{Hardware} Our approach was fully tested on an Intel(R) Core(TM) i9-10920X CPU @ 3.50GHz with a single 24 GB TESLA P40 GPU.

\subsection{Results}
\paragraph{\added{Accuracy in }continuous parsing} We first test all described linearizations under the proposed neural network on the continuous version of PTB. In Table~\ref{tab:con}, we report accuracies on dev and test splits and compare them against state-of-the-art approaches, including all existing sequence-to-sequence constituent parsers. While our approach achieves competitive accuracies with any tree linearization, we can observe that no substantial differences can be found between top-down and in-order strategies, and between enriched and regular variants (contrary to the observations by \cite{fernandez-gonzalez-gomez-rodriguez-2020-enriched} on RNN sequence-to-sequence constituent parsing). Moreover, we confirm that bottom-up linearizations underperform top-down and in-order variants also under this architecture, as in the results by \cite{Ma2017}, although the difference is smaller than in their case.

With respect to other sequence-to-sequence constituent parsers, our approach outperforms all existing models by a wide margin; and, in comparison with the best task-specific algorithms, top-down and in-order tree linearizations are only surpassed by models that are enhanced with the pre-trained language model XLNet \citep{XLNet}, notably larger than BERT \citep{devlin-etal-2019-bert} and RoBERTa. In fact, our approach is on par, for instance, with \cite{attachjuxtapose} (a purely transition-based parser) and \cite{tian-etal-2020-improving} (a chart-based model) when BERT$_\textsc{Large}$ is used instead.

\begin{table}[tbp]
\begin{center}
\centering
\begin{tabular}{@{\hskip 2pt}l@{\hskip 0pt}l@{\hskip 10pt}l@{\hskip 2pt}}
\toprule
\textbf{Parser} (no tags or predicted PoS tags) &  &\textbf{PTB} \\
\midrule
\citet{Liu2017} & & 91.8\hphantom{0} \\
\citet{stern-etal-2017-effective} & & 92.56 \\
\citet{fernandez-gonzalez-gomez-rodriguez-2018-dynamic} & & 92.0\hphantom{0} \\
\citet{fried-klein-2018-policy} & & 92.2\hphantom{0} \\
\citet{gaddy-etal-2018-whats} & & 92.08 \\
\citet{teng-zhang-2018-two} & & 92.4\hphantom{0} \\
\cite{Vilares_2020} + BERT$_\textsc{Large}$ & & 93.5\hphantom{0} \\
\citet{kitaev-etal-2019-multilingual} + BERT$_\textsc{Large}{}$ & & 95.59 \\
\citet{zhou-zhao-2019-head} + dependency + BERT$_\textsc{Large}{}$ & & 95.84 \\
\citet{zhou-zhao-2019-head} + dependency + XLNet & & 96.33 \\
\citet{mrini-etal-2020-rethinking} + dependency + POS + XLNet & & 96.38 \\
\cite{attachjuxtapose} + BERT$_\textsc{Large}$ & & 95.79 \\
\cite{attachjuxtapose} + XLNet & & 96.34 \\
\cite{tian-etal-2020-improving} + PoS + BERT$_\textsc{Large}$ & & 95.86 \\
\cite{tian-etal-2020-improving} + PoS + XLNet & & \textbf{96.40}\\
\cite{multipointer} + dependency + BERT$_\textsc{Large}{}$ & & 95.23\\ 
\hdashline[1pt/1pt]
\textit{\footnotesize (sequence-to-sequence models)} &   \\
\citet{Vinyals2015} &  & 88.3\hphantom{0}  \\
\citet{Vinyals2015} + ensemble &  & 90.5\hphantom{0}  \\
\citet{Vinyals2015} + ensemble + extra-data & & 92.8\hphantom{0}  \\
\citet{Ma2017} + ensemble & & 90.6\hphantom{0}  \\
\citet{Kamigaito2017} + ensemble & & 91.5\hphantom{0}  \\
\citet{Liu2018} + ensemble & & 92.3\hphantom{0}  \\
\citet{Suzuki2018} + ensemble + LM-rerank & & 94.32  \\
\citet{LiuS2S17} & & 90.5\hphantom{0}  \\
\cite{fernandez-gonzalez-gomez-rodriguez-2020-enriched} & & 91.6\hphantom{0}  \\
\cite{fernandez-gonzalez-gomez-rodriguez-2020-enriched} + \added{deterministic-attention} & & 91.2\hphantom{0}  \\
\cite{Vaswani2017} & & 91.3\hphantom{0} \\
\cite{Vaswani2017} + extra-data & & 92.7\hphantom{0} \\
\textbf{This work:} \hspace{8.5cm} &\textbf{(dev)} &\textbf{(test)}  \\
\ \ \ \ \textbf{SH-RE top-down linearization} & 95.56\tiny{$\pm$0.09} & 95.78\tiny{$\pm$0.06}\\
\ \ \ \ \textbf{enriched SH-RE top-down/bracketed linearization} & \textbf{95.63}\tiny{$\pm$0.07} & 95.70\tiny{$\pm$0.10} \\
\ \ \ \ \textbf{SH-RE in-order linearization} & 95.46\tiny{$\pm$0.03} & \textbf{95.84}\tiny{$\pm$0.02} \\
\ \ \ \ \textbf{enriched SH-RE in-order linearization} & 95.48\tiny{$\pm$0.03} & 95.71\tiny{$\pm$0.02} \\
\ \ \ \ \textbf{SH-RE bottom-up linearization} & 95.40\tiny{$\pm$0.15} & 95.58\tiny{$\pm$0.04} \\
\bottomrule
\end{tabular}
\centering
\setlength{\abovecaptionskip}{4pt}
\caption{F-score comparison of state-of-the-art constituent parsers on the PTB test split. The second block gathers exclusively sequence-to-sequence models. Parsers that use extra dependency information are marked with \textit{+dependency}, those that ensemble several trained models with \textit{+ensemble}, those that use a language model for reranking predicted trees with \textit{+LM-rerank}, those that use additional parsed data with \textit{+extra-data}, \added{those that use deterministic attention for increasing parsing speed with \textit{+deterministic-attention},} those that use predicted PoS tags as additional input with \textit{+PoS} and, finally, those that use pre-trained language models BERT$_\textsc{Large}$ \citep{devlin-etal-2019-bert} or XLNet \citep{XLNet} for the encoder initialization are marked with \textit{+BERT$_\textsc{Large}$/+XLNet}. We also include performance on the PTB dev split for all the tested linearizations. We report the average accuracy over 3 executions with different random seeds and standard deviations are indicated with $\pm$.}
\label{tab:con}
\end{center}
\end{table}

\paragraph{\added{Accuracy in }discontinuous parsing} We further evaluate the proposed sequence-to-sequence model and novel tree linearizations on dev (Table~\ref{tab:dev_disc}) and test (Table~\ref{tab:disc}) splits from discontinuous treebanks, additionally including in the latter table the best approaches to date for a comparison with the current state of the art. As observed on the continuous benchmark, top-down and in-order transition systems (augmented with the \textsc{Swap} transition) achieve similar overall F-scores on the English dataset; however, on German treebanks (especially  on NEGRA), the in-order tree linearization outperforms the top-down strategy. Regarding the accuracy on discontinuities, the top-down linearization obtains the best F-scores. Again, the bottom-up technique underperforms its counterparts in all datasets by a wide margin.

With respect to the alternatives with \textsc{Swap\#k} and \textsc{Shift\#k} tokens for shortening output sequences, we notice that the latter provides a poor performance and the former, while achieving a similar overall F-score to tree linearizations with the regular \textsc{Swap} action, 
shows a clear loss of accuracy
on discontinuities.

Overall,
our approach delivers competitive accuracies, outperforming recent task-specific discontinuous parsers (such as \citet{morbitz2020supertaggingbased} in TIGER and DPTB) and excelling in DPTB (where we achieve the best F-score and Discontinuous F-score to date).
 It can be also noticed that, 
 the sequence tagging strategy (enhanced with the attention mechanism provided by fully fine-tuning the language model BERT) by \citet{vilares-gomez-rodriguez-2020-discontinuous}, also included in Table~\ref{tab:con} as \cite{Vilares_2020} for the continuous version, is clearly outperformed 
 continuous and discontinuous benchmarks by our sequence-to-sequence model, which uses non-fine-tuned word embeddings.

\added{\paragraph{Parsing speed} We report in Table~\ref{tab:speed} the speeds provided by each linearization technique during decoding on the test splits. This comparison shows that the continuous and discontinuous bottom-up linearizations and the in-order variants with \textsc{Swap\#k} and \textsc{Shift\#k} transitions achieve the best speeds, with the latter being the fastest option in discontinuous constituent parsing. This behaviour was expected (and also empirically proved in the following section) since the target sequences generated by these linearizations are shorter than those produced by the other methods, with the in-order linearization with the \textsc{Shift\#k} token generating the shortest output sequences and being on par with the continuous in-order variant in speed (both 29 sent./s.). Although we leverage contextualized word embeddings in the proposed approach, any linearization of our model is twice as fast as other sequence-to-sequence methods in the continuous constituent benchmark, except for the model that applies a deterministic attention technique to speed up decoding \citep{fernandez-gonzalez-gomez-rodriguez-2020-enriched}. This latter system (which is based on the in-order linearization) is slightly faster than our more accurate variants (top-down and in-order), but it is surpassed by our model with the bottom-up linearization. Finally, while our implementation was not optimized for speed and the reported results are just intended for comparing the proposed linearization variants, we also include other discontinuous constituent parsers in the comparison, showing that our model is behind all of them\footnote{\added{Please note that \citet{fernandezgonzalez2021reducing} recently obtained such high speeds in discontinuous constituent parsing due to the application of faster continuous parsers after the original sentence was reordered by a pointer network \citep{Vinyals15}.}} and that there is still pending work in speeding up sequence-to-sequence models.} 

\begin{table*}[tbp]
\small
\centering
\begin{tabular}{@{\hskip 2pt}lccc@{\hskip 2pt}}
\toprule
\textbf{Tree linearization}  & \textbf{TIGER}
&\textbf{NEGRA}
&\textbf{DPTB}
\\
\midrule
\textbf{SH-RE top-down} + \textsc{Swap}   & \textbf{92.36}\tiny{$\pm$0.07} & 90.14\tiny{$\pm$0.03} & \textbf{95.44}\tiny{$\pm$0.03} \\
\textbf{SH-RE bottom-up} + \textsc{Swap}  & 91.48\tiny{$\pm$0.09} & 88.14\tiny{$\pm$0.02} & 94.81\tiny{$\pm$0.08} \\
\textbf{SH-RE in-order} + \textsc{Swap}  & 92.32\tiny{$\pm$0.02} & \textbf{90.79}\tiny{$\pm$0.08} & 95.32\tiny{$\pm$0.09} \\
\textbf{SH-RE in-order} + \textsc{Swap\#k}  & 92.27\tiny{$\pm$0.06} & 90.36\tiny{$\pm$0.14} & 95.25\tiny{$\pm$0.16} \\
\textbf{SH-RE in-order} + \textsc{Shift\#k}  & 91.11\tiny{$\pm$0.04} & 88.81\tiny{$\pm$0.16} & 94.76\tiny{$\pm$0.12} \\
\bottomrule
\end{tabular}
\centering
\setlength{\abovecaptionskip}{4pt}
\caption{
F-score on TIGER, NEGRA and DPTB development splits. We report the average accuracy over 3 executions with different random seeds and standard deviations are indicated with $\pm$.
}
\label{tab:dev_disc}
\end{table*}

\begin{table*}[tbp]
\small
\centering
\begin{tabular}{@{\hskip 2pt}l@{\hskip 6pt}c@{\hskip 6pt}cc@{\hskip 6pt}cc@{\hskip 6pt}c@{\hskip 2pt}}
\toprule
& \multicolumn{2}{c}{\textbf{TIGER}}
& \multicolumn{2}{c}{\textbf{NEGRA}}
& \multicolumn{2}{c}{\textbf{DPTB}}
\\
\midrule
\textbf{Parser} {\tiny (no tags or predicted PoS tags)} & \textbf{F1} & \textbf{DF1} & \textbf{F1} & \textbf{DF1} & \textbf{F1} & \textbf{DF1} \\
\hline
\citet{coavoux2019b} & 82.5 & 55.9  & 83.2 & 56.3 & 90.9 & 67.3  \\
\citet{coavoux2019a} & 82.7 & 55.9 & 83.2 & 54.6  & 91.0 & 71.3  \\
\citet{stanojevic-steedman-2020-span} & 83.4 & 53.5 & 83.6 & 50.7 & 90.5 & 67.1 \\
\citet{Corro2020SpanbasedDC} + BERT$_\textsc{X}$ & 90.0 & 62.1 & 91.6 & 66.1 & 94.8 & 68.9 \\

\citet{vilares-gomez-rodriguez-2020-discontinuous} + BERT$_\textsc{Base}$ & 84.6 & 51.1 & 83.9 & 45.6 & 91.9 & 50.8  \\
\citet{vilares-gomez-rodriguez-2020-discontinuous} + BERT$_\textsc{Large}$ & - & - & - & - & 92.8 & 53.9 \\
\citet{DiscoPointer} & 85.7 & 60.4 & 85.7 & 58.6 & - & -  \\
\citet{morbitz2020supertaggingbased} + BERT$_\textsc{Base}$ & 88.3 & 69.0 & 90.9 & 72.6 & 93.3 & 80.5 \\
\citet{fernandezgonzalez2021reducing} + BERT$_\textsc{Base}$ & 88.5 & 63.0 & 90.0 & 65.9 & 94.0 & 68.9  \\
\citet{fernandezgonzalez2021reducing} + BERT$_\textsc{Large}$ & \textbf{90.5} & 68.1 & \textbf{92.0} & 67.9 & 94.7 & 72.9 \\
\citet{fernandezgonzalez2021reducing} +  XLNet & - & - & - & - & \textbf{95.1} & \textbf{74.1}  \\
\citet{multipointer}+ dep + BERT$_\textsc{Base}$ & 89.8 & \textbf{71.0} & 91.0 & \textbf{76.6} & - & -  \\
\hdashline[1pt/1pt]
\textbf{This work:} & \tiny{$\pm$0.05} & \tiny{$\pm$0.27} & \tiny{$\pm$0.09} & \tiny{$\pm$1.35} & \tiny{$\pm$0.06} & \tiny{$\pm$0.46}\\
\ \ \ \ \textbf{SH-RE top-down} + \textsc{Swap} \textbf{ linearization}  & 88.28 & \textbf{67.95} & 88.59 & \textbf{67.43} & 95.37 & \textbf{83.85}\\
& \tiny{$\pm$0.14} & \tiny{$\pm$0.61} & \tiny{$\pm$0.19} & \tiny{$\pm$0.65} & \tiny{$\pm$0.04} & \tiny{$\pm$0.49}\\
\ \ \ \ \textbf{SH-RE bottom-up} + \textsc{Swap} \textbf{ linearization} & 87.02 & 63.20 & 85.74 & 57.76 & 95.12 & 82.40 \\
& \tiny{$\pm$0.04} & \tiny{$\pm$0.45} & \tiny{$\pm$0.02} & \tiny{$\pm$0.84} & \tiny{$\pm$0.06} & \tiny{$\pm$0.25}\\
\ \ \ \ \textbf{SH-RE in-order} + \textsc{Swap} \textbf{ linearization} & \textbf{88.53} & 67.76 & \textbf{89.08} & 67.06 & 95.47 & 83.80 \\
& \tiny{$\pm$0.08} & \tiny{$\pm$0.49} & \tiny{$\pm$0.10} & \tiny{$\pm$0.76} & \tiny{$\pm$0.01} & \tiny{$\pm$0.38}\\
\ \ \ \ \textbf{SH-RE in-order} + \textsc{Swap\#k} \textbf{ linearization} & 88.36 & 65.68 & 88.93 & 65.38 & \textbf{95.48} & 82.86 \\
& \tiny{$\pm$0.12} & \tiny{$\pm$0.16} & \tiny{$\pm$0.13} & \tiny{$\pm$0.41} & \tiny{$\pm$0.09} & \tiny{$\pm$0.66}\\
\ \ \ \ \textbf{SH-RE in-order} + \textsc{Shift\#k} \textbf{ linearization} & 87.10 & 54.27 & 86.76 & 46.86 & 94.96 & 69.17 \\
\bottomrule
\end{tabular}
\centering
\setlength{\abovecaptionskip}{4pt}
\caption{
F-score and Discontinuous F-score (DF1) comparison of state-of-the-art discontinuous constituent parsers on TIGER, NEGRA and DPTB test splits. Parsers that use extra dependency information are marked with \textit{+dep}, 
and those that use pre-trained language models BERT$_\textsc{base}$, BERT$_\textsc{Large}$ \citep{devlin-etal-2019-bert} or XLNet \citep{XLNet} for the encoder initialization are marked with \textit{+BERT$_\textsc{base}$/+BERT$_\textsc{Large}$/+XLNet} (we use +BERT$_\textsc{X}$ when the model size was not specified). We report the average accuracy over 3 executions with different random seeds and standard deviations are indicated with $\pm$.
}
\label{tab:disc}
\end{table*}

\begin{table*}[tbp]
\small
\centering
\begin{tabular}{@{\hskip 2pt}l@{\hskip 6pt}c@{\hskip 6pt}c@{\hskip 6pt}c@{\hskip 6pt}c@{\hskip 2pt}}
\toprule
\textbf{Parser}& {\textbf{TIGER}}
& {\textbf{NEGRA}}
& {\textbf{DPTB}}
& {\textbf{PTB}}
\\
\midrule
\citet{zhou-zhao-2019-head} + XLNet & - & - & - & 65\\
\citet{mrini-etal-2020-rethinking} + XLNet & - & - & - & 59\\
\cite{attachjuxtapose} + XLNet & - & - & - & 71\\
\citet{vilares-gomez-rodriguez-2020-discontinuous} + BERT$_\textsc{Base}$ & 80 & 80 & 80 & - \\
\citet{vilares-gomez-rodriguez-2020-discontinuous} + BERT$_\textsc{Large}$ & - & - & 34  & - \\
\citet{morbitz2020supertaggingbased} + BERT$_\textsc{Base}$ & 60 & 68 & 57 & - \\
\citet{fernandezgonzalez2021reducing} + BERT$_\textsc{Base}$ &  \textbf{238} & \textbf{275}  & \textbf{231} & -  \\
\citet{fernandezgonzalez2021reducing} + BERT$_\textsc{Large}$ & 207 & 216 & 193 & - \\
\citet{fernandezgonzalez2021reducing} +  XLNet & - & - & 179 & -\\
\hdashline[1pt/1pt]
\textit{\footnotesize (sequence-to-sequence models)} &   \\
\citet{LiuS2S17}  & - & - & - & 17\\
\cite{fernandez-gonzalez-gomez-rodriguez-2020-enriched}  & - & - & - & 17\\
\cite{fernandez-gonzalez-gomez-rodriguez-2020-enriched} + deterministic-attention  & - & - & - & 35\\
\textbf{This work:} & & & \\
\ \ \ \ \textbf{SH-RE top-down} \textbf{ linearization}  & - & - & - & 29\\
\ \ \ \ \textbf{SH-RE bottom-up} \textbf{ linearization}  & - & - & - & \textbf{39}\\
\ \ \ \ \textbf{SH-RE in-order} \textbf{ linearization} & - & - & - & 29\\
\ \ \ \ \textbf{SH-RE top-down} + \textsc{Swap} \textbf{ linearization}  & 26 & 21 & 21 & -\\
\ \ \ \ \textbf{SH-RE bottom-up} + \textsc{Swap} \textbf{ linearization}  & 29 & 27 & 26 & -\\
\ \ \ \ \textbf{SH-RE in-order} + \textsc{Swap} \textbf{ linearization} & 26 & 21 & 21 & -\\
\ \ \ \ \textbf{SH-RE in-order} + \textsc{Swap\#k} \textbf{ linearization}  & 30 & 29 & 24 & -\\
\ \ \ \ \textbf{SH-RE in-order} + \textsc{Shift\#k} \textbf{ linearization}  & \textbf{47} & \textbf{50} & \textbf{29} & -  \\
\bottomrule
\end{tabular}
\centering
\setlength{\abovecaptionskip}{4pt}
\caption{
\added{Speed comparison (sentences/second) of our approach with different linearization techniques on TIGER, NEGRA, DPTB and continuous PTB test splits. We also add in the second block those sequence-to-sequence models whose speed can be found in the literature, including the approach by \cite{fernandez-gonzalez-gomez-rodriguez-2020-enriched} that speeds up parsing decoding with deterministic attention (marked with \textit{+deterministic-attention}). In addition, 
we present in the first block top-performing constituent parsers augmented with pre-trained language models BERT$_\textsc{base}$, BERT$_\textsc{Large}$ \citep{devlin-etal-2019-bert} or XLNet \citep{XLNet} (\textit{+BERT$_\textsc{base}$/+BERT$_\textsc{Large}$/+XLNet}). Please note that the reported speeds from previous work were measured on different hardware setups.}
}
\label{tab:speed}
\end{table*}

\begin{table*}[tbp]
\small
\centering
\begin{tabular}{@{\hskip 0pt}l@{\hskip 6pt}c@{\hskip 6pt}cc@{\hskip 6pt}cc@{\hskip 6pt}cc@{\hskip 6pt}c@{\hskip 0pt}}
\toprule
& \multicolumn{2}{c}{\textbf{TIGER}}
& \multicolumn{2}{c}{\textbf{NEGRA}}
& \multicolumn{2}{c}{\textbf{DPTB}}
& \multicolumn{2}{c}{\textbf{PTB}}
\\
\midrule
\textbf{Tree linearization} & \textbf{Size} & \textbf{Length} & \textbf{Size} & \textbf{Length} & \textbf{Size} & \textbf{Length} & \textbf{Size} & \textbf{Length} \\
\hline
\textbf{SH-RE top-down}  & - & - & - & - & - & - & 29 & 367\\
\textbf{enriched SH-RE top-down/bracketed}  & - & - & - & - & - & - & 55 & 367\\
\textbf{SH-RE in-order} & - & - & - & - & - & - & 30 & 368\\
\textbf{enriched SH-RE in-order} & - & - & - & - & - & - & 56 & 368\\
\textbf{SH-RE bottom-up} & - & - & - & - & - & - & 202 & 255 \\
\textbf{SH-RE top-down} + \textsc{Swap}   & 28 & 2067 & 30 & 2061 & 31 & 1497 & - & -\\
\textbf{SH-RE bottom-up} + \textsc{Swap} & 191 & 2039 & 205 & 2021 & 204 & 1444  & - & - \\
\textbf{SH-RE in-order} + \textsc{Swap}  & 29 & 2068 & 31 & 2062 & 32 & 1498 & - & -\\
\textbf{SH-RE in-order} + \textsc{Swap\#k}  & 66 & 1150 & 63 & 1202 & 57 & 873 & - & -\\
\textbf{SH-RE in-order} + \textsc{Shift\#k}  & 66 & 257 & 63 & 208 & 57 & 368 & - & -\\
\bottomrule
\end{tabular}
\centering
\setlength{\abovecaptionskip}{4pt}
\caption{
Output dictionary size (Size) and length of the longest target sequence (Length) in TIGER, NEGRA, DPTB and continuous PTB training datasets for each proposed linearization technique.}
\label{tab:info}
\end{table*}

\subsection{Analysis}
\label{sec:analysis}
It can be argued that a large output dictionary size harms performance, at least, this is one of the reasons that might explain the difference in accuracy between our approach and sequence labeling techniques, where the target vocabulary is significantly larger to keep synchronicity between the length of the input and output sequences. However, some studies, such as \cite{fernandez-gonzalez-gomez-rodriguez-2020-enriched}, claim that, by properly augmenting (\textit{enriching}) the vocabulary, accuracy gains can be obtained. 

In order to better understand why some tree linearizations are underperforming others \added{in terms of accuracy} (more notable in discontinuous parsing), we 
report in Table~\ref{tab:info} the target vocabulary size and longest sequences for each tree linearization and treebank. From this information, we can extract that:
\begin{itemize}
    \item The two best-performing linearizations (regular top-down and in-order methods) present the smallest output dictionaries both in continuous and discontinuous datasets.
    \item Enriched variants in PTB almost duplicate output dictionary sizes, but obtain scores on par with the regular versions (not providing substantial accuracy gains as reported for RNN sequence-to-sequence models \citep{fernandez-gonzalez-gomez-rodriguez-2020-enriched}).
    \item The bottom-up strategy requires a larger vocabulary (due to the parameterized \textsc{Reduce\#k-X}) that might penalize its performance. Note that this would hold even on other bottom-up transition systems that require a previous binarization, since this transformation enlarges the amount of non-terminal labels and, as a consequence, the output vocabulary.
    \item Dealing with short target sequences does not lead to a better performance, since the use of the token \textsc{Shift\#k} generates the shortest output sequences in discontinuous treebanks, but underperforms practically all other methods.
    \item Finally, it can be also observed that, although target sequences in discontinuous datasets are significantly longer than those in PTB, the attention mechanism avoids deteriorating accuracy on long sequences, achieving similar accuracies, for instance, on PTB and DPTB. 
    
\end{itemize}

\noindent Therefore, 
the correlation between vocabulary size 
and performance can be 
observed,
but the target sequence length 
seems to have
no impact \added{in accuracy} thanks to the attention mechanism,
which is the likely reason why \textsc{Swap\#k} and \textsc{Shift\#k} seem to produce no discernible advantage over the \textsc{Swap} transition under our neural architecture.

\added{Moreover, we can clearly see in Table~\ref{tab:info} a correlation between target sequence length and decoding speed, regardless of dictionary size. The higher speeds in continuous and discontinuous constituent parsing are respectively delivered by the bottom-up and the in-order+\textsc{Shift\#k} linearizations (as shown in Table~\ref{tab:speed}), which clearly are the variants that produce the shortest target sequences. Even the fact that the bottom-up alternative has the largest dictionary among continuous linearizations seems to have no effect in decoding speed. In addition, the continuous in-order variant generates output sequences with the same length as those provided by the in-order+\textsc{Shift\#k}, obtaining the same speed during decoding time. Finally, we can also observe that the top-down and in-order linearizations (with and without the \textsc{Swap} token augmentation) are the slowest options since they generate the longest target sequences.} 

We also believe that the logic of each transition system and how the output tokens encode the resulting constituent tree can affect the text-to-parse translation. For instance, from a transition-based standpoint, the use of the \textsc{Shift\#k} action allows the model to operate over the whole buffer by just predicting the \textsc{Shift} transition parameterized with the correct $k$ value, covering a broader context and losing the locality typically found in classic transition-based algorithms (which exclusively modify the first word in the buffer and/or the two words on top of the stack). In sequence-to-sequence models, this means that the model has a broader search space, not only due to dealing with a larger output dictionary, but also due to having several options for the same purpose: \new{in this case, different \textsc{Shift\#k} tokens can be used for representing each input word, instead of using a single \textsc{Shift} transition}. This might explain why the linearization with \textsc{Shift\#k} underperforms the other alternatives that behave like classic shift-reduce algorithms and have a single \textsc{Shift} token to read words from the input.  
\new{Moreover, this same reasoning can be used for explaining how the availability of different \textsc{Swap\#k} tokens (instead of applying a single \textsc{Swap}) 
penalizes accuracy prediction and leads to a worse performance on discontinuities.} 
On the other hand, transition systems that mark the beginning and the end of each constituent with \textsc{Non-Terminal-X} and \textsc{Reduce} tokens (as can be seen in sequences generated by the top-down and in-order algorithms) tend to work better as tree linearizations than the bottom-up strategy (which denotes the span of each constituent with a single \textsc{Reduce\#k-X} at the end). This might be the main explanation why bottom-up approaches are less adequate for encoding phrase structure trees.

\begin{figure}
\begin{center}
\includegraphics[width=\columnwidth]{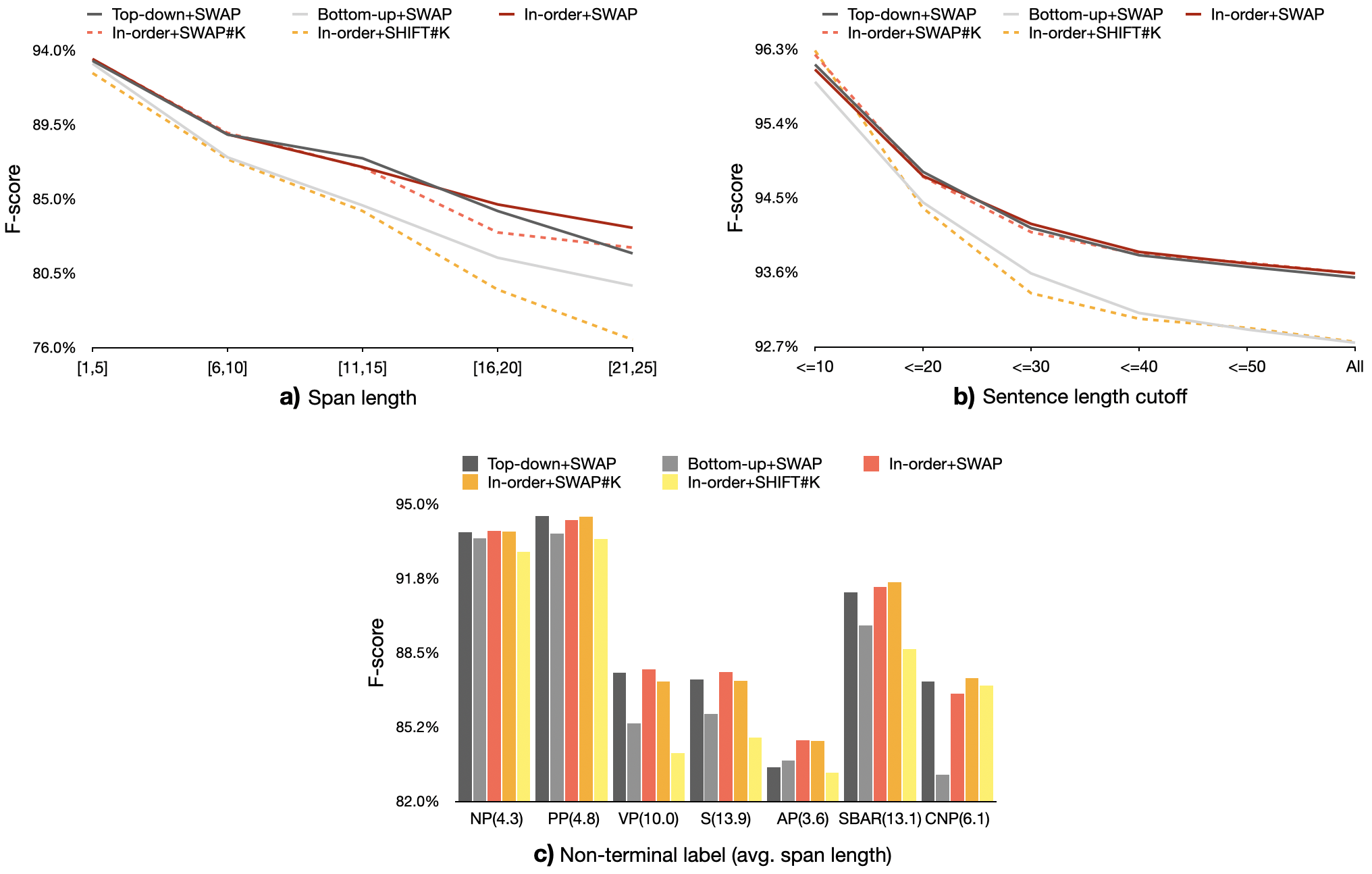}
\end{center}
\caption{Accuracy of discontinuous linearization methods relative to structural factors and sentence length.}
\label{fig:analysis}
\end{figure}

Additionally, to provide more evidence that can help us understand the differences in performance between different linearizations, we undertake an error analysis relative to structural factors and sentence lengths 
on a concatenation of the dev splits from the three discontinuous treebanks. In particular, 
Figure~\ref{fig:analysis}(a) shows the F-score on span identification for different lengths, Figure~\ref{fig:analysis}(b) presents the performance on different sentence length cutoffs and Figure~\ref{fig:analysis}(c) plots the accuracy when assigning the most frequent non-terminal labels (including the average span length in brackets) by each proposed discontinuous tree linearization. From that information, we can claim that:
\begin{itemize}
    \item Error propagation, 
    often observed
    in purely transition-based parsers, can be also seen in Figure~\ref{fig:analysis}(a) and (b) for sequence-to-sequence models.
    As expected,
    sequential prediction 
    can cause
    earlier mistakes 
    to
    affect future decisions,
    resulting in 
    more errors in the encoded constituent tree.
    Its impact can be 
    seen on long spans and long sentences, where the accuracy of all linearizations decreases.
    \item While the in-order+\textsc{Swap} linearization shows the best performance on phrases with longer spans; using \textsc{Swap\#k} instead harms accuracy, especially when the span length increases (probably due to the fact that a larger amount of \textsc{Swap\#k} are required for reordering a longer sequence of input words, and more mistakes can be made during that process).
    \item The in-order variant with the \textsc{Shift\#k} transition suffers notable accuracy loses on producing longer constituents, probably because it is more likely to make a mistake and predict the wrong \textsc{Shift\#k} token in constituents that cover more input words (represented by \textsc{Shift\#k} tokens in the tree linearization) and, therefore, the impact of error propagation is higher. A similar trend can be seen for the bottom-up strategy as, while \textsc{Reduce\#k-X} tokens with a lower $k$ value are more frequent and easier to learn, wrong predictions are more likely on longer constituents.
    \item Based on the performance on longer sentences, we can also note that error propagation has a higher impact on bottom-up and in-order+\textsc{Shift\#k} linearizations than on the other alternatives. 
    \item No significant differences in performance between the top-down+\textsc{Swap} and in-order+\textsc{Swap} strategies can be found in Figure~\ref{fig:analysis}(a) and (b); \new{however, in Figure~\ref{fig:analysis}(c), we can see that the top-down approach is 
    substantially 
    outperformed by the in-order strategy on building constituents of type AP (Adjective Phrase), even being slightly surpassed by the bottom-up strategy on that task. While no frequent patterns were observed in the data to explain that behaviour, a lower precision on this kind of structures (i.e., tagging as AP constituents of a different type) with respect to the other linearizations is the reason of these differences in F-score.
    } 
    \item Finally, the bottom-up linearization method has important drops in accuracy on the creation of constituents of type VP, S, SBAR and, especially, CNP. While the low performance on constituents VP, S and SBAR can be explained by the fact that they have a high span length and this linearization is more prone to suffer from error propagation; \new{the poor accuracy on constituents of type CNP (Coordinated Noun Phrases) is caused by failing to correctly identify the boundaries of that kind of phrases when long enumerations of Noun Phrases (separated by commas) have to be processed.}
    The variant with the \textsc{Shift\#k} transition also has a poor performance on large constituents VP, S and SBAR; but surprisingly outperforms the in-order+\textsc{Swap} (one of the best-performing linearizations) on building CNP constituents.
\end{itemize}


\section{Conclusions}
\label{sec:conclusion}
In this article, we present the first sequence-to-sequence constituent parser that can produce discontinuous phrase structure trees. To achieve that, we define novel transition systems for linearizing discontinuous structures and present a more powerful neural architecture to implement a state-of-the-art sequence-to-sequence model. The resulting system not only accurately produces discontinuous constituent trees, but also achieves the best accuracy to date among sequence-to-sequence constituent parsers on the main benchmarks, and 
advances the state of the art in accuracy
on DPTB.

Finally, it is worth mentioning that, to the best of our knowledge, neither of the novel transition systems defined in this work have been studied in a purely transition-based framework and, since the in-order algorithm \citep{Liu2017} achieves the best accuracy to date for a transition-based parser on continuous treebanks, the variant enhanced with the \textsc{Swap} action might outperform current state-of-the-art models on discontinuous benchmarks.

\section*{Acknowledgments}
We acknowledge the European Research Council (ERC), which has funded this research under the European Union’s Horizon 2020 research and innovation programme (FASTPARSE, grant agreement No 714150) and the Horizon Europe research and innovation programme (SALSA, grant agreement No 101100615), ERDF/MICINN-AEI (SCANNER-UDC, PID2020-113230RB-C21), Xunta de Galicia (ED431C 2020/11), and Centro de Investigaci\'on de Galicia ``CITIC'', funded by Xunta de Galicia and the European Union (ERDF - Galicia 2014-2020 Program), by grant ED431G 2019/01.








\printcredits

\bibliographystyle{cas-model2-names}

\bibliography{anthology,main}


\clearpage



\end{document}